\documentclass[sigconf]{acmart}

\usepackage{graphicx}
\usepackage[many]{tcolorbox}
\usepackage{mathtools,amssymb,latexsym,amsfonts,stmaryrd}
\usepackage{pbox}
\usepackage{amsthm}
\usepackage{multirow}
\usepackage{tikz}
\usepackage{mathrsfs}
\usepackage{mathpartir}
\usepackage[ruled,linesnumbered]{algorithm2e}
\usepackage{url}
\usepackage{syntax}
\usepackage{framed}
\usepackage[noend]{algpseudocode}
\usepackage{flushend}
\usepackage{listings}
\usepackage{moresize}
\usepackage{wrapfig}
 
\usepackage{seqsplit}
\usepackage{alltt}
\usepackage{xspace}
\usepackage{enumitem}
\usepackage{pifont}

\DeclareMathAlphabet{\mathcal}{OMS}{cmsy}{m}{n}

\usepackage{amsmath, listings, amsthm, amssymb, proof, xspace}

\usepackage{thmtools}

\declaretheoremstyle[spaceabove=\topsep,notefont=\normalfont\itshape]{mystyle}

\newcommand{\revise}[2]{{\color{red}{\ifx&#1&\else- #1\fi}} {\color{ForestGreen}{\ifx&#2&\else+ #2\fi}}}%
\renewcommand{\revise}[2]{#2}%

\usetikzlibrary{arrows,patterns, decorations.pathreplacing}

\newcommand{\F}{Fig.}
\newcommand{\T}{Table}
\renewcommand{\S}{Sec.}

\newcommand{\ignore}[1]{}

\lstdefinestyle{base}{
  moredelim=**[is][\color{red}]{@}{@},
  escapeinside={<@}{@>}
}

\newcommand{\tool}{\textsc{MetaOD}\xspace}

\newcommand{\synbracket}[1]{[\![#1]\!]}

\usepackage{amssymb}
\usepackage{pifont}

\newcommand\DejaVuttfamily{%
  \fontfamily{DejaVuSansMono-TLF}\selectfont
}

\lstdefinestyle{base}{
  moredelim=**[is][\color{red}]{@}{@},
  escapeinside={<@}{@>}
}

\lstdefinelanguage
   [x64]{Assembler}     
   [x86masm]{Assembler} 
   {morekeywords={CDQE,CQO,CMPSQ,CMPXCHG16B,JRCXZ,LODSQ,MOVSXD, %
                  POPFQ,PUSHFQ,SCASQ,STOSQ,IRETQ,RDTSCP,SWAPGS, %
                  rax,rdx,rcx,rbx,rsi,rdi,rsp,rbp, %
                  r8,r8d,r8w,r8b,r9,r9d,r9w,r9b}} 

\usepackage[compact]{titlesec}

\usepackage{listings}
\usepackage{fancyvrb}
\usepackage{color}
\definecolor{lightgray}{rgb}{.9,.9,.9}
\definecolor{darkgray}{rgb}{.4,.4,.4}
\definecolor{purple}{rgb}{0.65, 0.12, 0.82}
\definecolor{commentgreen}{RGB}{63,127,95}

\usepackage{xcolor}
\colorlet{myPurple}{blue!40!red}
\definecolor{myOrange}{RGB}{255,192,0}

 \lstdefinelanguage{Solidity}{
   keywords={len,delete,int,void,payable, public, event, contract, typeof, new, true, false, catch, function, return, null, catch, switch, var, if, in, while, do, else, case, break,unsigned,int32_t,int16_t,for},
   keywordstyle=\color{violet}\bfseries,
   ndkeywords={class, export, boolean, throw, implements, import, this},
   ndkeywordstyle=\color{darkgray}\bfseries,
   identifierstyle=\color{black},
   sensitive=false,
   comment=[l]{//},
   morecomment=[s]{/*}{*/},
   commentstyle=\color{commentgreen}\ttfamily,
   stringstyle=\color{red}\ttfamily,
   morestring=[b]',
   morestring=[b]"
 }

\newcommand{\rnum}[1]{\uppercase\expandafter{\romannumeral #1\relax}}

\usepackage{xcolor}

\algnewcommand{\LeftComment}[1]{\Statex \(\triangleright\) #1}

\definecolor{pptbrown}{RGB}{132,60,12}
\definecolor{pptgreen}{RGB}{56,87,35}

\AtBeginDocument{%
  \providecommand\BibTeX{{%
    \normalfont B\kern-0.5em{\scshape i\kern-0.25em b}\kern-0.8em\TeX}}}

\setcopyright{acmcopyright}
\copyrightyear{2018}
\acmYear{2018}
\acmDOI{10.1145/1122445.1122456}

\acmConference[]{}{}{}
\acmPrice{15.00}
\acmISBN{978-1-4503-9999-9/18/06}

\setcopyright{none}
\renewcommand\footnotetextcopyrightpermission[1]{} 

\begin{document}

\title{Metamorphic Testing for Object Detection Systems}

\author{Shuai Wang}
\affiliation{%
  \institution{The Hong Kong University of Science and Technology}
}

\email{shuaiw@cse.ust.hk}

\author{Zhendong Su}
\affiliation{%
  \institution{ETH Zurich}
}
\email{zhendong.su@inf.ethz.ch}


\begin{abstract}
Recent advances in deep neural networks (DNNs) have led to object detectors that
can rapidly process pictures or videos, and recognize the objects that they
contain. Despite the promising progress by industrial manufacturers such as
Amazon and Google in commercializing deep learning-based object detection as a
standard computer vision service, object detection systems --- similar to
traditional software --- may still produce incorrect results. These errors, in
turn, can lead to severe negative outcomes for the users of these object
detection systems. For instance, an autonomous driving system that fails to
detect pedestrians can cause accidents or even fatalities. However, principled,
systematic methods for testing object detection systems do not yet exist,
despite their importance.

To fill this critical gap, we introduce the design and realization of \tool, the
first metamorphic testing system for object detectors to effectively reveal
erroneous detection results by commercial object detectors. To this end, we (1)
synthesize natural-looking images by inserting extra object instances into
background images, and (2) design metamorphic conditions asserting the
equivalence of object detection results between the original and synthetic
images after excluding the prediction results on the inserted objects. \tool\ is
designed as a streamlined workflow that performs object extraction, selection,
and insertion. We develop a set of practical techniques to realize an effective
workflow, and generate diverse, natural-looking images for testing.
Evaluated on four commercial object detection services and four pretrained
models provided by the TensorFlow API, \tool\ found tens of thousands of
detection defects in these object detectors. To further demonstrate the
practical usage of \tool, we use the synthetic images that cause erroneous 
detection results to retrain the model. Our results show that the
model performance is increased significantly, from an mAP score of 9.3 to an
mAP score of 10.5. 
\end{abstract}

%


\settopmatter{printfolios=true}

\maketitle

\vspace{-1pt}
\section{Introduction}
\label{sec:introduction}
Deep learning-based object detectors identify objects in a given image using
convolutional neural networks. Currently, several major industrial
manufacturers, including Google, Amazon, Microsoft, and Lockheed Martin, are
building and improving object detectors to serve as the basis for various
computer vision tasks. These models are widely-used in real-world applications,
such as optical character recognition (OCR), ball tracking in sports, pedestrian
detection systems in autonomous cars, robotics, and machine inspection. They are
also used as an initial step in surveillance and medical image analysis
applications, which often require highly precise and reliable detection results.

Despite this spectacular progress, however, deep learning-based object
detection systems --- similar to traditional software --- can yield erroneous
prediction results that are potentially disastrous. In particular, given the
widespread adoption of object detection systems in critical applications in the
security, medical, and autonomous driving fields, incorrect or unexpected
edge-case behaviors have caused severe threats to public safety or financial
loss~\cite{tian2018deeptest,abdessalem2018testing,damon2018uber}. For
instance, in one infamous case in 2016, Tesla's autopilot mode caused a fatal
crash when the autonomous driving system failed to recognize a white truck
against a bright sky~\cite{neal2017tesla}. More recently, an Uber autonomous driving
system killed a pedestrian crossing the road, which is believed to have
been due to the system's failure in recognizing a pedestrian in dark
clothing~\cite{david2018uber}.

\begin{figure*}[!ht]
  \centering \includegraphics[width=0.99\linewidth]{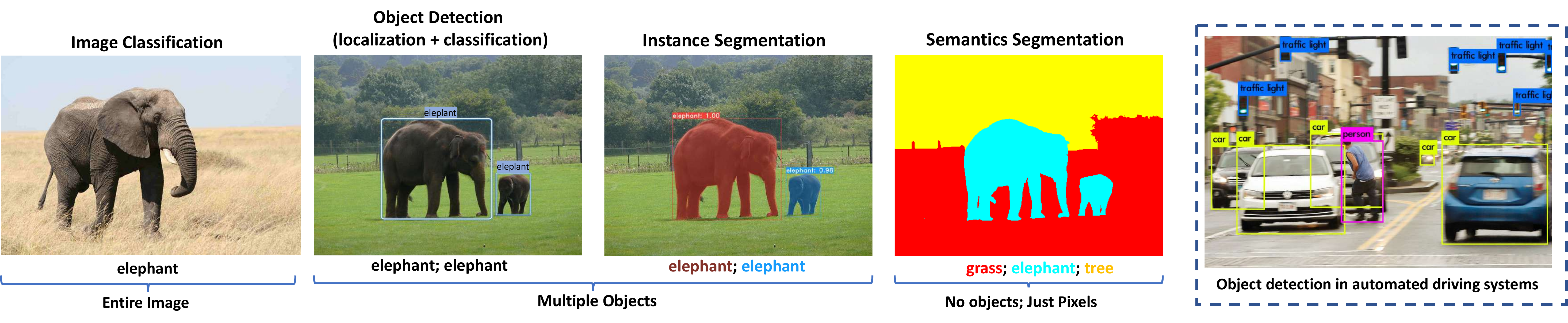}
  \vspace*{-5pt}
  \caption{Typical image analysis tasks solved by deep learning techniques and
    sample usage of object detection for traffic scenes (the last figure).}
  \label{fig:cv}
  \vspace*{-5pt}
\end{figure*}

In recent years, a number of techniques have been designed to test deep learning
systems, such as convolutional neural networks (CNN) and recurrent neural
networks (RNN) models~\cite{pei2017deep,du2019deep}. The techniques have also
been applied to test domain-specific applications such as autodriving
systems~\cite{tian2018deeptest,abdessalem2018testing,zhang2018deeproad} and to
test the underlying infrastructure of deep learning
libraries~\cite{pham2019cradle,kim2019guiding,dutta2018testing}. However, the
principles specific for testing object detection systems have not been investigated
by existing research, which, thus, unlikely results in comprehensive, systematic
testing of object detection systems.

This paper tackles this important problem by introducing
the first metamorphic testing~\cite{chen1998metamorphic,chen2018mtr} technique, \tool, aiming at 
effectively exposing erroneous prediction results of commercial object detection
systems. Given a real image as the ``background'', \tool\ inserts an object
instance into the background, generates a synthetic image, and then employs a
metamorphic condition to check the consistency of object detection results
between the synthetic image and the corresponding background.
To effectively generate \textit{diverse} and \textit{natural-looking} images
that trigger \textit{practical} prediction errors, \tool\ is designed as a
three-step approach, performing object extraction, object refinement/selection,
and object insertion. The object extraction module extracts object instance
images from a large set of pictures using advanced instance segmentation
techniques~\cite{bolya2019yolact}, thus aggregating many object sets
distinguished by category. Then, given a background image, the object
refinement/selection module implements a set of lightweight albeit effective
criteria for selecting certain objects from object sets that are closely related
to the background. To determine insertion locations, the object insertion module
uses domain-specific criteria and techniques enlightened by delta
debugging~\cite{zeller1999yesterday} to find locations that presumably trigger
prediction errors, while retaining realism and diversity of the synthetic images
to a good extent.

The proposed workflow shows promising abilities and findings; we evaluated four
commercial object detection services provided by Amazon, Google, IBM, and
Microsoft~\cite{amazon2018rekognition,google2018api,microsoft2019api,ibm2019api}
and four pretrained models provided by the TensorFlow object detection
API~\cite{tensorflow2018sdk}. Our testing revealed tens of thousands of
erroneous object detection results from these commercial services. In addition,
we retrained an object detection model using synthetic images that cause this
model to output erroneous outputs. The evaluation results show that the model
performance improved substantially after retraining. In summary, this work
makes the following main contributions:

\begin{itemize}
\item We introduce a novel metamorphic testing approach for object
  detection systems, vital components in various computer vision
  applications (e.g., self-driving cars). Our technique treats object detectors
  as ``black-boxes''. Thus, it is highly generalizable for testing
  real-world object detectors, such as remote services on the cloud.
\item To generate diverse and natural-looking sets of images as the test inputs,
  we design and realize \tool, a streamlined workflow that performs object extraction,
  object refinement/selection, and object insertion to synthesize input images
  in an efficient and adaptive manner.
\item Our approach tests object detectors in a realistic setting and delineates
  the capabilities of state-of-the-art commercial object detectors. From a total of
  292,206 input images, \tool\ found 38,345 erroneous detection results in eight
  popular (commercial) object detectors. By leveraging synthetic images that
  trigger erroneous object detector outputs for retraining, we show that the
  performance of object detection models can be substantially improved.
\end{itemize}
%

\section{Background}
\label{sec:background}

\subsection{Deep-Learning for Image Analysis}
\label{subsec:image-analysis}
Deep learning has achieved substantial success in various challenging computer
vision problems. \F~\ref{fig:cv} reviews four typical tasks that deep learning
techniques address well.\footnote{Images used in writing this manuscript were
  collected using Google image search with non-restricted usage rights.}
Indeed, these tasks are the basis of many computer vision applications,
including image captioning, dense captioning, and object tracking.

Image classification is a fundamental task that attempts to comprehend an entire
image as a whole. The goal is to classify the image by assigning it to a
specific label. Typically, image classification refers to images in which only
one object appears and is analyzed. In contrast, object detection involves both
classification and localization tasks, and is used to analyze more realistic
cases in which multiple objects may exist in an image. Object detection attempts
to recognize the objects in an input image by categorizing each object and
determining appropriate bounding boxes for the identified objects. As shown in
the last image in \F~\ref{fig:cv}, modern object detection, enabled by deep
learning techniques, has become a key technique used by autonomous driving
systems to recognize traffic lights, signs, pedestrians and other vehicles on
the roads. Technical details of object detection techniques are given in
\S~\ref{subsec:object-detection}.

Advanced computer vision tasks, instance
segmentation~\cite{romera2016recurrent,uhrig2018box2pix,bolya2019yolact}, are
intended to achieve finer-grained object localization in input images. The
bounding boxes used in object detection find only coarse-grained object
boundaries and include many pixels that do not belong to the object. In
contrast, instance segmentation improves the object localization accuracy by
identifying each pixel that acts as part of a known object in the image. The
semantic segmentation task~\cite{noh2015learning,li2017fully} involves
associating each pixel in an image with a class label. This line of
research aims to enable complete scene understanding of images, and is still a
developing line of research in the field of computer vision. To date, both
instance and semantic segmentation techniques have been applied to industrial
inspection and medical imaging analysis tasks.

\subsection{Object Detection}
\label{subsec:object-detection}

Object detection was conventionally addressed using handcrafted features and
selective region
searches~\cite{uijlings2013selective,perronnin2010improving,wang2009hog}. The
input images are dissected into small regions (each region is called a ``region
proposal'' and is likely to contain an object) via
heuristics~\cite{uijlings2013selective}. Then, features are extracted from each
region proposal for object classification. To date, two major lines of research
(popular models proposed in both line of research are tested in this work; see
\S~\ref{sec:evaluation}) exist that have drastically improved object detection
techniques with deep learning, both of which are briefly introduced below.

\begin{figure}[t]
  \centering \includegraphics[width=0.95\linewidth]{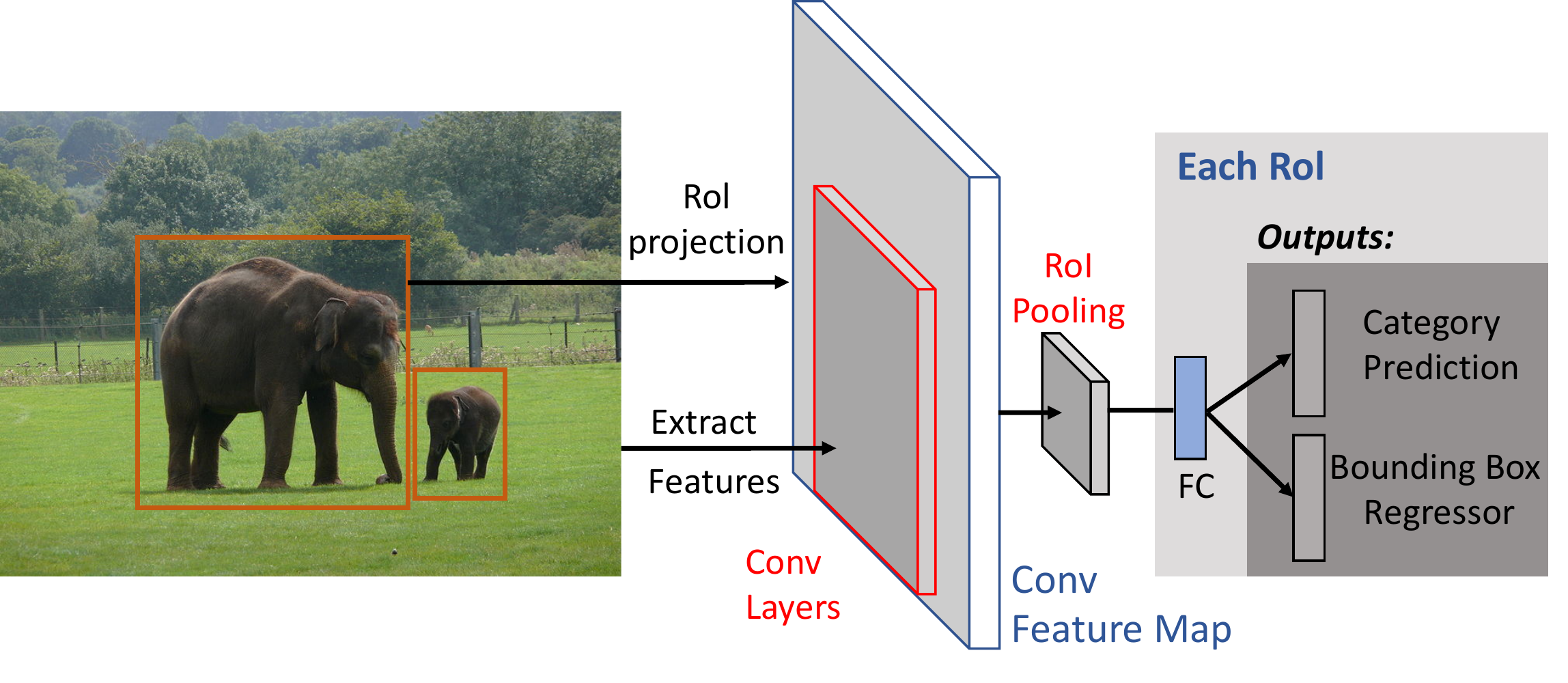}
  \caption{A simplified Fast RCNN workflow illustrating typical two-stage object
    detector architectures.}
  \label{fig:rcnn}
  \vspace*{-5pt}
\end{figure}

\noindent \textbf{Two-Stage Region-Based Object Detectors.}~Motivated by the
primary success in applying deep neural networks for image
classification~\cite{krizhevsky2017imagenet}, RCNN~\cite{roos2014rich} was among
the first to apply convolutional neural networks (CNN) for object detection.
The proposed technique forms a two-stage pipeline in which each region proposal
extracted from the input image is an input to CNN for feature extraction.  Then,
the extracted features are forwarded to an SVM classifier and a bounding box
regressor to determine the object category and bounding box offsets,
respectively.
%
Since then, object-detection research has focused on rapidly evolving the RCNN
architecture~\cite{ren2015faster,dai2016r,he2016deep} and removing explicit
dependence on region proposals to improve speed.
Fast-RCNN~\cite{ross2015fastrcnn} introduced a modern \textit{end-to-end}
prediction pipeline. As shown in \F~\ref{fig:rcnn}, instead of region proposals,
the entire image is forwarded to the CNN to generate a convolutional feature map
and region proposals are extracted from the feature map (\textit{first stage}).
A Region of Interest (RoI) pooling layer is placed before the fully connected
layer (FC) to reshape each proposal into a fixed size, and FC layer's outputs
are fed to softmax and bbox regressor layers for object classification and for
determining bounding box offsets, respectively (\textit{second stage}).
%
%


\begin{figure}[t]
  \centering \includegraphics[width=0.95\linewidth]{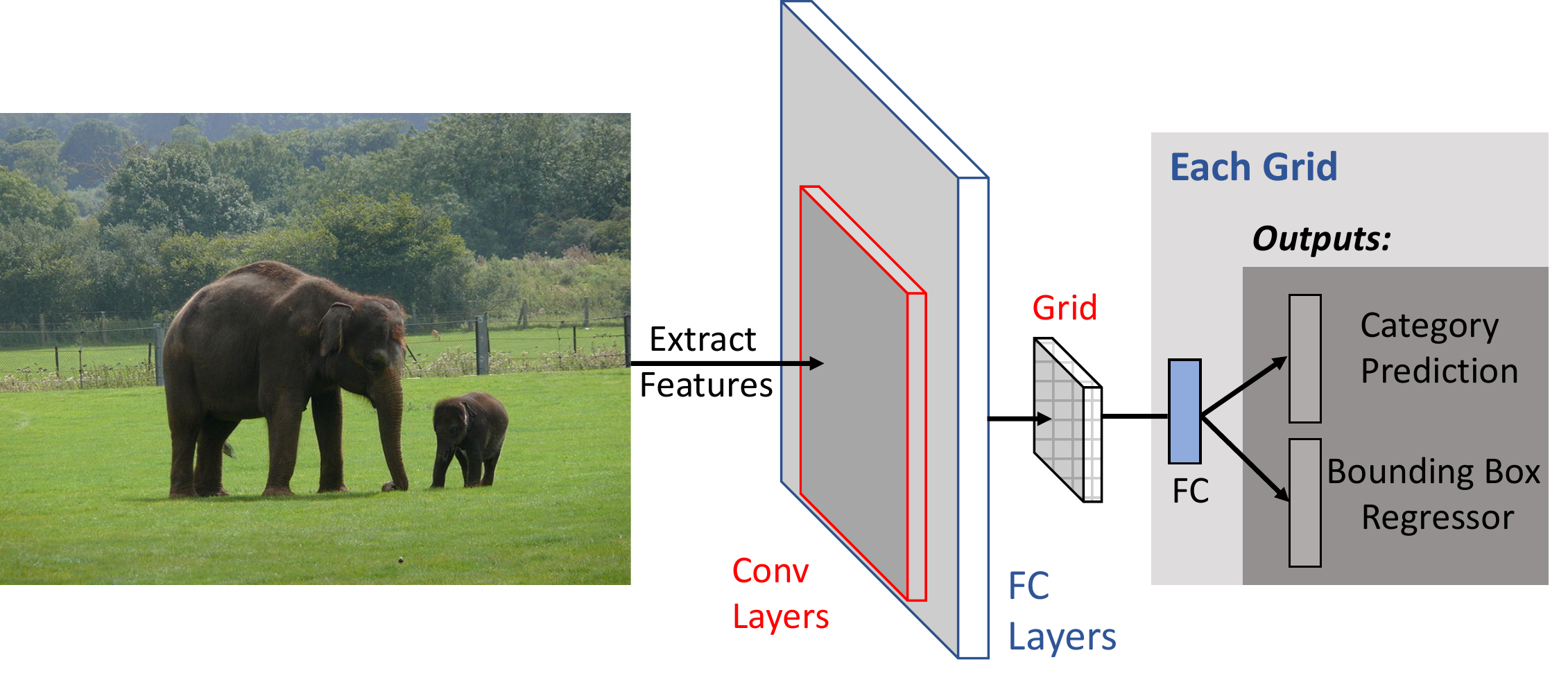}
  \caption{A simplified YOLO workflow illustrating typical single-stage object
    detector architectures.}
  \label{fig:yolo}
  \vspace*{-5pt}
\end{figure}

\noindent \textbf{Single-Stage Object Detectors.}~Two-stage object detectors use
regions, explicitly or implicitly, for object localization. Another line of
research aims to propose a cost-effective solution without region proposals by
designing a single-stage feed-forward CNN network in a monolithic setting. Such
networks are usually less computationally intensive by trading precision for
speed, and are usually more suitable for real-time tasks or for use in mobile
devices.

The YOLO (You Only Look
Once)~\cite{redmon2015yolo,redmon2016yolo,redmon2018yolo} and SSD (Single Shot
Detector)~\cite{liu2016ssd} models are de facto object detectors that feature
single-stage architectures. \F~\ref{fig:yolo} depicts the YOLO workflow, in
which input images are first divided into an $S \times S$ grid; then, a fixed
number of bounding boxes are predicted within each grid. For each bounding box,
the network outputs a class probability and the bounding box offsets. A bounding
box is deemed to contain objects when its class probability exceeds a threshold
value. The entire pipeline is typically orders of magnitude faster than
region-based techniques. Indeed, the object extraction module of \tool\ is built
on top of YOLACT~\cite{bolya2019yolact}, a real-time instance segmentation model
that was inspired by the YOLO object detection model.


\begin{figure*}[t]
  \centering
  \includegraphics[width=1.0\linewidth]{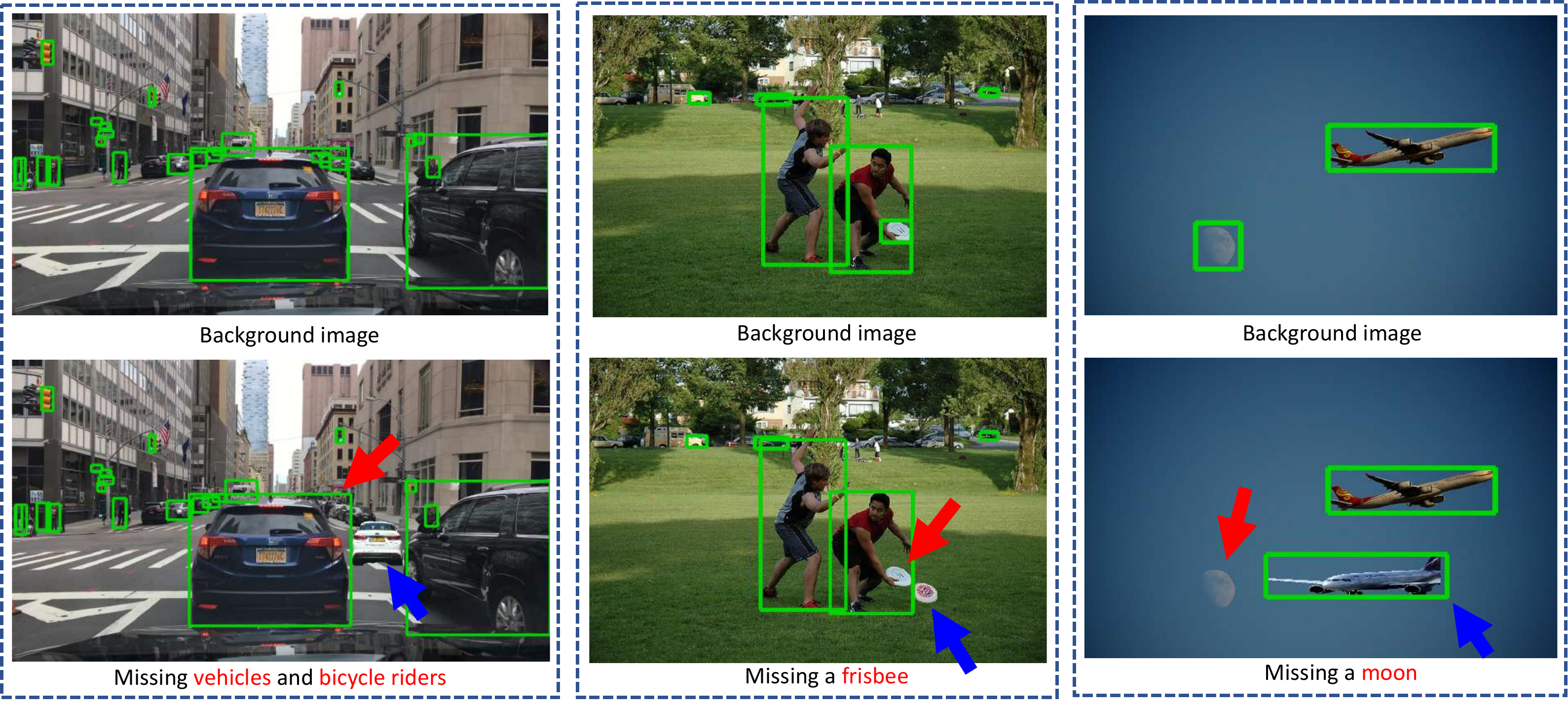}
  \vspace*{-5pt}
  \caption{Object detection errors found by \tool. We slightly cherry picked
    images in favor of readability. Browse the full results at~\cite{dropbox}.
    The inserted objects are pointed by \textcolor{blue}{blue arrows}. To
    preserve realism of the synthetic images at our best effort, the inserted
    objects are resized to the average size of existing objects of the same
    category. See our discussions in \S~\ref{subsec:design-refinement}.}
  \label{fig:case}
\end{figure*}

\section{Approach Overview}
\label{sec:approach-overview}
Metamorphic testing (MT) has been widely used to automatically generate tests to
detect software faults~\cite{chen1998metamorphic,chen2018mtr}. The strength of
MT lies in its capability to alleviate the test oracle problem via metamorphic
relations (MRs). Each MR depicts necessary properties of the target software in
terms of inputs and their expected outputs. In other words, even if the
correctness of actual outputs are difficult to determine, it is possible to construct
and check proper MRs among the expected outputs of the given inputs to detect
software detects. In this research, we apply metamorphic testing to object
detectors. To provide an overview of our approach, we start by formulating the
relevant notations.

By feeding a test image $i$ to an object detector $d$, the prediction output is
denoted as $d\synbracket{i}$, which consists of $N$ three-element tuples
$(b_{k}, l_{k}, c_{k})$, where $N$ denotes the number of objects recognized in
$i$, $b_{k}$ the location of the $k$th object recognized in $i$, $l_{k}$
the category label, and $c_{k}$ the confidence score of the prediction.
Then, given a set of object instance images $\mathbb{O}$ extracted from a large
number of real images (see \F~\ref{subsec:design-extraction}), and a set
$\mathbb{C}$ where each $\mathcal{C} \in \mathbb{C}$ is a 2-D coordinate $(x,y)$ in
$i$, a synthetic image $i'$ can be represented as:

$$i' = \mathcal{C}(o, i), o \in \mathbb{O}~\text{and}~\mathcal{C} \in \mathbb{C}$$

\noindent where $o$ is placed such that its cendroid is at the 2-D coordinate
specified by $\mathcal{C}$. Note that in this research, we do \textbf{not} apply
any transformation rules (rotation, blurring, etc.) on the inserted objects to
preserve realism at our best effort, and $\mathcal{C} \in \mathbb{C}$ is
deliberately constructed such that the inserted object $o$ does \textbf{not}
overlap with preexisting objects in the ``background'' image $i$. Therefore, the
MR adopted in this research can be formalized as follows:

\[
\forall \mathcal{C} \in \mathbb{C}\; \forall o \in \mathbb{O}.\; \mathcal{E}(d\synbracket{i}, d\synbracket{i'}-(b_{o},l_{o},c_{o})) = \mathit{True}
\]

\noindent where $i' = \mathcal{C}(o, i)$. Here, we \textbf{exclude} the
prediction result on $o$ (i.e., tuple $(b_{o}, l_{o}, c_{o})$) from
$d\synbracket{i'}$, and $\mathcal{E}$ is a criterion asserting the equality of
object detection results (details will be given shortly in
\S~\ref{subsec:equal}). The given MR is defined such that no matter how the
image $i'$ is synthesized by inserting an additional object $o$ on $i$, the
object detection results are expected to be consistent with those in the
original image. Consequently, erroneous predictions can be revealed by checking
the failure of the given MR.

While the given MR holds for any synthetic image $i' = \mathcal{C}(o, i)$, one
practical problem is that not all the synthetic images represent real-world
scenarios. Indeed, there exists research in the CV community where
unrealistic images are synthesized to train image analysis models, for example,
by placing a car on the table~\cite{tremblay2018training}. While the synthetic
``unrealistic'' images may fulfill the requirement of model training in previous
work, we aim to also augment the realism of the synthetic images such that flagged
erroneous behaviors unveil practical defects that can cause confusion during
daily usage of object detectors. Additionally, as we will explain in
\S~\ref{subsec:design-insertion}, randomly deciding a position for insertion
without considering preexisting objects' positions in the background image $i$
would undermine the effectiveness of the proposed technique.

Therefore, in this research, we gather $\mathbb{O'} \subset \mathbb{O}$ such
that $\mathbb{O'}$ contains object instance images that are closely related to
the background image $i$ (see \S~\ref{subsec:design-refinement}). We also form 
favorable insertion locations $\mathbb{C'} \subset \mathbb{C}$ likely
to trigger prediction errors by leveraging empirical evidence and strategies
enlightened by delta debugging (see \S~\ref{subsec:design-insertion}). Hence,
the MR is modified as follows:

\[
   \forall \mathcal{C} \in \mathbb{C'}\; \forall o \in \mathbb{O'}.\; \mathcal{E}(d\synbracket{i}, d\synbracket{i'}-(b_{o},l_{o},c_{o})) = \mathit{True}
\]


\begin{figure}[!ht]
  \centering \includegraphics[width=0.90\linewidth]{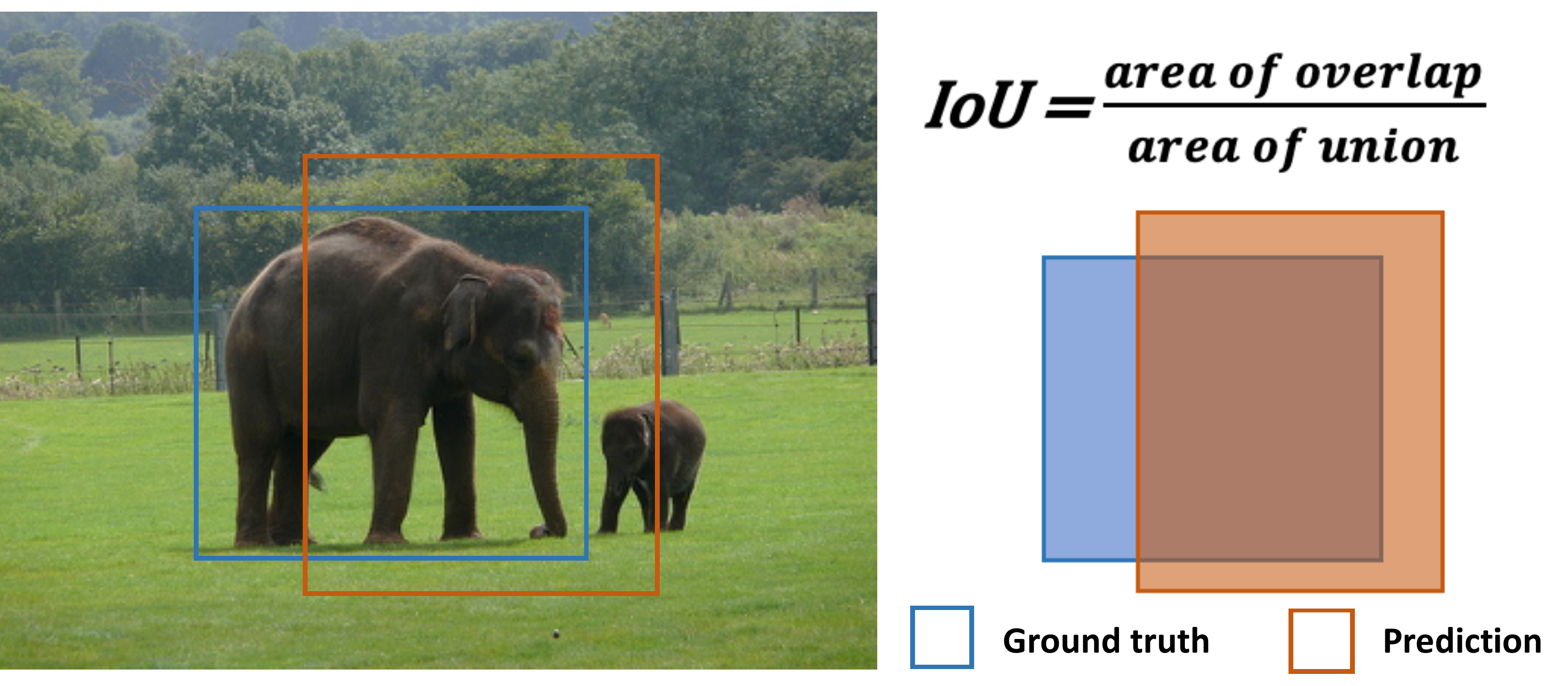}
  \caption{Intersection over Union (IoU).}
  \vspace*{-7pt}
  \label{fig:iou}
\end{figure}

\subsection{Equality Criteria}
\label{subsec:equal}
Asserting the \textit{equality} of object detection outputs (i.e., N
three-element tuples $(b_{k}, l_{k}, c_{k})$) is indeed too strict because
bounding boxes of certain objects could be slightly drifted within each round of
prediction. The CV community instead uses a standard metric, Average
Precision (AP)~\cite{everingham2010pascal}, to compensate small localization
drifting when evaluating object detector accuracy. Note that the AP score is
computed by taking both ``precision'' and ``recall'' values into account, as we
will explain later in this section. In this research, our equality criteria
$\mathcal{E}$ is derived from the AP score.

To compute AP, Intersection over Union (IoU) is used to measure each object
detection boundary with respect to the ground truth. As shown in
\F~\ref{fig:iou}, IoU measures the overlap between two bounding boxes with the
same prediction label (i.e., ``elephant''), and denotes how much the
predicted boundary overlaps with the ground truth. In case IoU is greater than a
threshold $\epsilon$ (e.g., 0.5), the prediction is deemed a \textit{true
  positive}. The precision and recall scores are then computed by taking all the
prediction results into account, and the AP score can be further derived by
computing the area under the precision-recall
curve~\cite{kuznetsova2018open,russakovsky2015imagenet}. For an image with
objects of different categories, the mean AP (mAP) is computed by averaging all
AP scores. In our setting, the prediction results of the ``background'' image $i$
entails the ground truth, and are compared with detection results of the
synthetic image $i' = \mathcal{C}(o, i)$. Since $o$ does not overlap with
existing objects on $i$, and, therefore, does not interfere with relevant predictions,
the mAP score is expected to be 100\%. Thus, $\mathcal{E}$ is defined as
follows:

\begin{displaymath}
\begin{array}{l}
\hspace{-6pt} \mathcal{E}(d\synbracket{i}, d\synbracket{i'}-(b_{o},l_{o},c_{o})) \doteq \\
  mAP(d\synbracket{i}, d\synbracket{i'}-(b_{o},l_{o},c_{o})) = 100\%
\end{array}
\end{displaymath}

To date, multiple variants of the standard mAP definition exist.  We adopt one of
the most popular mAP calculation methods, the PASCAL VOC
metric~\cite{kuznetsova2018open}, for our implementation.
%

\begin{figure*}[t]
  \vspace*{-5pt}
  \centering \includegraphics[width=0.85\linewidth]{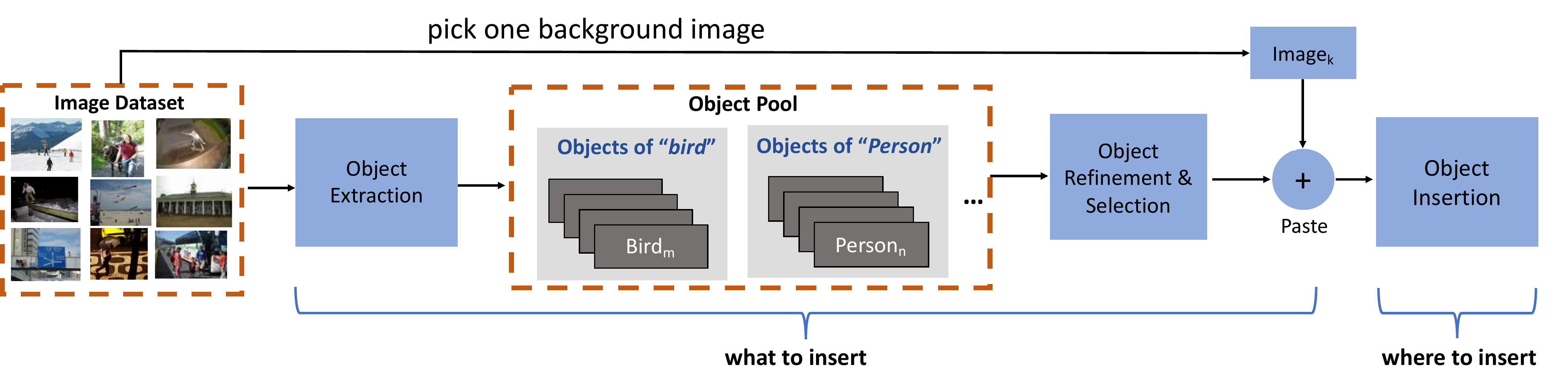}
  \vspace*{-5pt}
  \caption{Workflow of \tool.}
  \label{fig:workflow}
\end{figure*}

\subsection{Case Study}
\label{subsec:case-study}
The evaluation criterion $\mathcal{E}$ defined in \S~\ref{subsec:equal} enables a
unified approach to check object detection failures --- it is image
content agnostic and therefore can be automatically conducted. From a holistic
perspective, the following categories of object detection defects can be obtained by
checking $\mathcal{E}$:

\begin{itemize}
  \item \textbf{Recognition failures} represent errors which treat an arbitrary
    region on the image containing no object as an ``object'' or fail to
    recognize an existing object.
  \item \textbf{Classification failures} represent labeling errors, for instance
    labeling a human being as a ``bird.''
  \item \textbf{Localization failures} represent the failure where the object
    detector uses a too large or too small bounding box to localize objects. As
    illustrated in \F~\ref{fig:iou}, too large driftings ($IoU < \epsilon$) on
    the bounding box are not allowed.
\end{itemize}

Nevertheless, after manually checking object detection failures found by \tool,
we only find \textit{recognition failures}.\footnote{We found over 28K images
  triggering object detection errors (see \S~\ref{sec:evaluation}). We manually
  checked about 800 images by re-querying the remote services with the error
  triggering images and screened the detection outputs.} \tool\ has successfully
found a large number of object detection failures by eight popular (commercial)
object detection services (see \T~\ref{tab:model}). \F~\ref{fig:case} reports
three cases, where the ``background'' images on the first row are from the
Berkeley DeepDrive dataset~\cite{bdd,yu2019bdd} and the COCO
dataset~\cite{lin2014microsoft}. Images on the second row are generated by
inserting one extra object on their corresponding ``background.''

By inserting extra objects (indicated by the \textcolor{blue}{blue arrows}) into
the background and checking the equality criteria $\mathcal{E}$
(\S~\ref{subsec:equal}), we were able to provoke many detection defects. The
first column in \F~\ref{fig:case} illustrates a recognition failure (indicated
by the \textcolor{red}{red arrows}), where a bike rider and several cars in the
traffic scene image could not be recognized after a new vehicle was inserted.
Similarly, the synthetic images in the second and third columns unveil detection
failures, where after inserting one extra object into images of real-world
scenes, existing objects (moon and frisbee) cannot be recognized. We note that
\F~\ref{fig:case} demonstrates the \textit{diversity} in the issues we found;
\tool\ synthesizes test images of different scenes and therefore can find a
broad set of defects. In contrast, existing relevant
works~\cite{tian2018deeptest,zhang2018deeproad} are primarily designed to
transform or synthesize images of only driving scenes (see
\S~\ref{subsec:application-scope}).



\subsection{Application Scope}
\label{subsec:application-scope}
It is worth noting that we are \textit{not} testing extreme cases to stress the
object detection systems~\cite{alcorn2019strike}. Apparently, we can synthesize
images that are highly challenging to human beings and therefore challenging to
object detectors as well, for instance, by tweaking the contrast of objects and
its background. Additionally, unnatural blending of a pasted object with its
background will affect the prediction~\cite{dwibedi2017cut}. Therefore, while in
this research we propose a set of techniques to select ``realistic'' objects for
insertion (see \S~\ref{subsec:design-refinement} and
\S~\ref{subsec:design-insertion}), we still define a conservative test oracle
such that we \textit{exclude} the prediction over the newly inserted object, and
check only the consistency of the remaining predictions.

Existing approaches~\cite{tian2018deeptest,zhang2018deeproad} apply predefined
``severe weather conditions'' (e.g., foggy and rainy) to transform or directly
synthesize \textit{entire images}. They are not tailored to pinpoint object
detection failures, and are conceptually orthogonal to the
object-level mutations proposed in this work. In addition, their transformations
may be inapplicable to mutate arbitrary images while preserving realism. For
instance, applying severe weather conditions toward images of indoor scenes is
likely unreasonable.

\section{Design}
\label{sec:design}
\F~\ref{fig:workflow} depicts a holistic view of the proposed technique. To
generate image $i' = \mathcal{C}(o, i)$ for testing, \tool\ is constructed as a
streamlined workflow that includes object extraction, object
selection/refinement, and object insertion modules. By providing \tool\ with a
set of images (e.g., images from the COCO dataset~\cite{lin2014microsoft}), its
object extraction module performs advanced object instance segmentation
techniques to identify object instances in a set of images
(\S~\ref{subsec:design-extraction}). Then, given an image as the ``background'',
the object selection module determines an appropriate object to be inserted in
the background (\S~\ref{subsec:design-refinement}), using a set of criteria to
find similar objects, rule out low-quality objects and adjust the object size.
While the first two steps address the challenge of ``what to insert'', for a
particular background image, we need to further answer the question of ``where
to insert.'' We aggregate empirical evidence and derive heuristics to select
insertion positions. Furthermore, motivated by how delta
debugging~\cite{zeller1999yesterday} is applied to test conventional software,
we propose techniques to augment the diversity of synthetic images
(\S~\ref{subsec:design-insertion}).

\subsection{Object Extraction}
\label{subsec:design-extraction}
The first step in our streamlined process is object extraction, which is
performed to extract a pool of objects from input images. As mentioned in
\S~\ref{subsec:image-analysis}, while object extraction is generally considered
difficult, deep learning-enabled instance segmentation has been shown to work
well in practice~\cite{bolya2019yolact,li2017fully,he2017mask}. Therefore, in
this study, we reuse existing instance segmentation techniques to collect object
instance images from natural images.

Similar to research conducted for object detection, instance segmentation also
has two primary focuses: accuracy~\cite{li2017fully,he2017mask} and
speed~\cite{uhrig2018box2pix,bolya2019yolact}.
%
%
In this work, we concentrate on models that emphasize speed over performance.
The object extraction module is designed to swiftly extract objects from large
sets of diverse images. Therefore, speed takes priority over accuracy (although
in practice our adopted instance segmentation model has a good accuracy as
well). In \S~\ref{subsec:design-refinement}, we compensate for the ``accuracy''
of extracted objects by proposing techniques to rule out low-quality object
images. Overall, by orchestrating object extraction and refinement modules in a
streamlined workflow, we output sets of high-quality labeled object images with
a modest cost and high speed.

To this end, we reuse YOLACT~\cite{yolact}, a recently developed real-time
instance segmentation tool, to build the object extraction module. Our empirical
evidence (also reported in its accompanying paper~\cite{bolya2019yolact}) shows
that YOLACT has impressive speed and quite good accuracy in practice when
processing real-world images. YOLACT outputs a mask over each recognized object
instance (see \F~\ref{fig:cv}). We extend YOLACT by reusing the object masks to
extract each object from the background images. Therefore, when we feed the object
extraction model with an image, for example the second ``elephant'' image in
\F~\ref{fig:cv}, the output of this step is a set of two object images, each of
which is labeled an ``elephant.''

\subsection{Object Refinement and Selection}
\label{subsec:design-refinement}
Despite significant progress, instance segmentation remains a difficult problem,
and we have observed that some of its outputs are of low quality. According to our
observations, these ``low-quality'' object images occur for two main reasons: (1)
some objects in the input image are too small, and (2) some objects overlap and
therefore fragmentary object images are extracted.\footnote{Such general
  challenges still exist even if we tentatively tried more ``heavy-weight''
  instance segmentation models like the Tensorflow implementation of Mask
  RCNN~\cite{he2017mask}.} We acknowledge the general difficulty of outputting
high-quality object images. Instead, our object extraction module processes
large sets of images at high speed, and we further prune low-quality objects and
select appropriate objects closely related to a ``background'' image.

\noindent \textbf{Small Object Image Pruning.}~As shown in
\F~\ref{fig:workflow}, the output of object extraction consists of multiple sets
of images, where each set contains object instances with the same label. During this
step, we first prune small object images within each object set, which
presumably include low-resolution or fragmentary images unsuitable for
use. To perform pruning, we sort the object instance images within
each set by image size and remove the majority of object images (in our 
implementation, we remove 90\% of the object images).

\noindent \textbf{Object Image Similarity Analysis.}~For a particular
``background'' image with several preexisting objects, we aim to find object
instance images from the pool that are closely related to the background to
fulfill the requirement of testing the object detector while also preserving the
realism of the synthetic images as much as possible. To this end, we perform an
image similarity analysis using image hashing techniques. Image hashing is a
standard technique for pixel-level image similarity analysis. The process
creates similar hashes for similar images. In contrast, when using a crypto hash
algorithm such as MD5, one byte of difference can lead to drastic hash value
changes due to the avalanche effect~\cite{patil2016comprehensive}.

Given an image $i$ with three ``birds'', we start by computing the average image
hash value of these ``bird'' object images. Then we iterate over all the ``bird''
images in the pool (see \F~\ref{fig:workflow}) and identify a ``bird'' whose
image hash value has the shortest Hamming distance with the average hash value.
This ``bird'' will be used for insertion. If image $i$ contains objects with $N$
different labels, we repeat the procedure $N$ times. Therefore, $N$ objects of
different categories will be selected for insertion. In this way, we ensure the
``realism'' of the synthetic images as much as possible. Our observations show
that the selected ``similar'' object images can usually exhibit texture and
resolution that are close to those of the background image.

For the implementation, we use the average hash~\cite{ahash}, which is a
standard implementation for image hashing. Our tentative tests showed that this
method helps find similar objects to a good extent at modest cost.
Nevertheless, we acknowledge the difficulty, if it is at all possible, of
finding \textit{semantically similar} objects through a unified and
cost-efficient approach. Indeed, image hashing uses \textit{pixel-level}
similarity instead of reflecting on the \textit{meaning} of each object
instance. We leave for future work the exploration of practical techniques to
comprehend the semantic information of each object instance and refine object
selection at this step.

\noindent \textbf{Object Image Resizing.}~Before inserting a selected object
image into a background image $i$, we adjust the object size to match that of
the existing objects in image $i$. We resize the object image to the average
size of the objects in $i$ that belong to the same category. Also, as notated in
\S~\ref{sec:approach-overview}, besides adaptively resizing, we do not
``transform'' object images (rotation, blurring, etc.) to preserve realism at
our best effort.

\begin{figure}[!ht]
  \centering \includegraphics[width=0.80\linewidth]{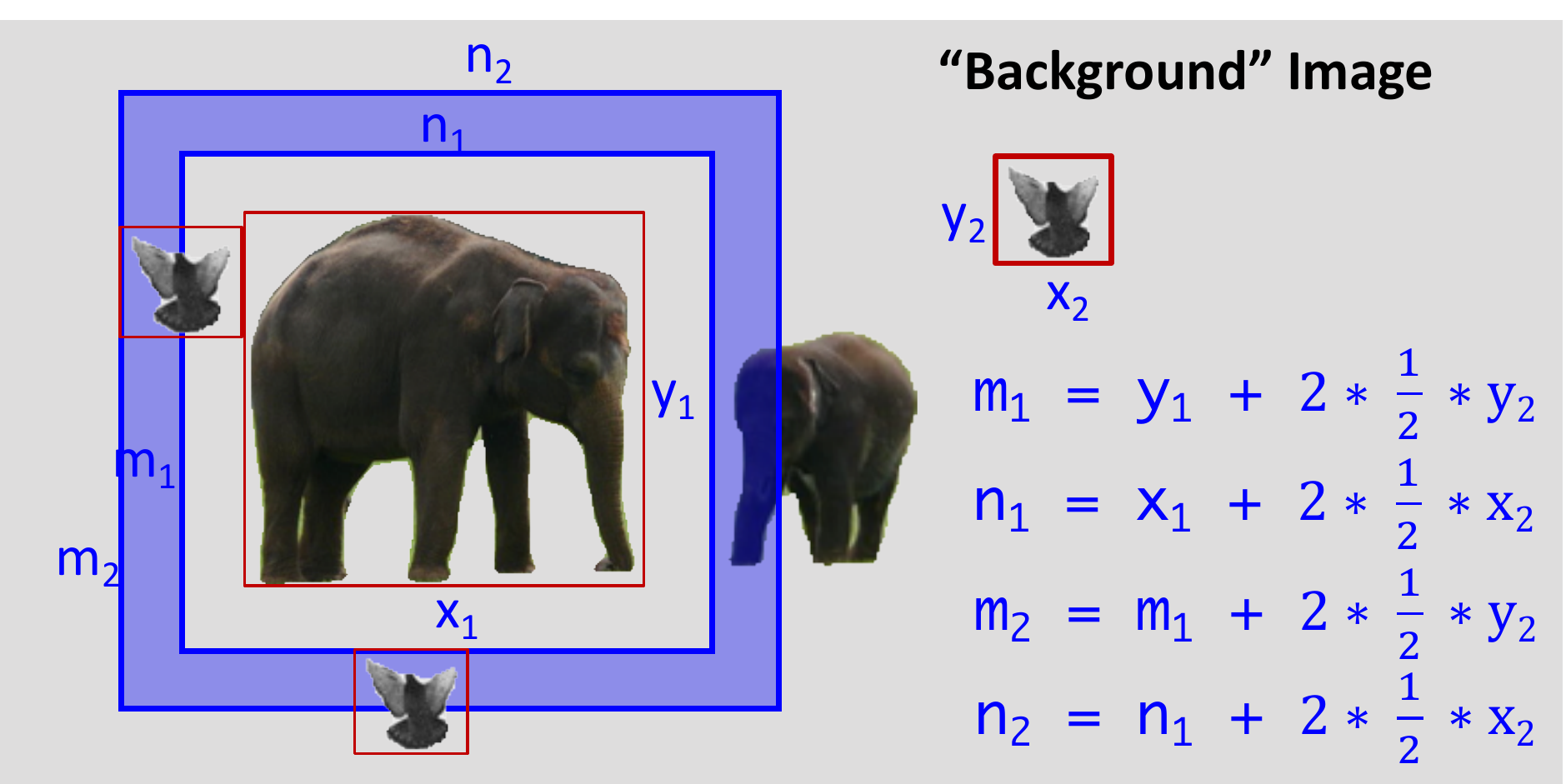}
  \caption{The ``guided insertion'' strategy to insert a ``bird.'' The
    \textcolor{blue}{blue region} is symmetrical and centered on the larger
    ``elephant.''}
  \label{fig:guided-insertion}
\end{figure}

\subsection{Object Insertion}
\label{subsec:design-insertion}
After selecting proper objects, we then seek proper locations on the background
image for insertion. As discussed in \S~\ref{subsec:application-scope}, the
software engineering (SE) community transforms entire images for testing, while the
computer vision (CV) community primarily concerns with the visual appearance of
the inserted object, rather than the ``background'' into which the object is
placed~\cite{dwibedi2017cut,tremblay2018training,hinterstoisser2018pre}. Several
studies have attempted to infer reasonable insertion locations using statistical
methods such as probabilistic grammar models and have only applied them to
images of indoor scenes~\cite{qi2018human}. However, building a generalized
model for arbitrary scenes, if at all possible, is highly challenging in this
research, where large-scale synthetic images are required to reveal erroneous
object detection results.

Given the general difficulty of leveraging heavy-weight statistical methods to
infer ``optimal'' insertion locations, we instead propose lightweight
strategies. In this section, we start by conducting empirical studies on
locations where insertion can presumably trigger object detection defects. Then,
motivated by delta debugging used in testing traditional
software~\cite{zeller1999yesterday}, we augment the diversity of the synthetic
images by progressively relocating the inserted objects on the background
images.

\noindent \textbf{Determining Object Insertion Locations.}~Our preliminary
studies show that inserting objects \textit{close to} existing objects in an
image (referred to as \textit{guided insertion} later in this paper) is likely
to trigger erroneous predictions. This section presents empirical results
to support our observation. To set up the study, we randomly selected 50 images
from the COCO image set~\cite{lin2014microsoft} and tentatively inserted a
``bird'' image. As reported in this section, we tested eight popular object
detection models and show the evaluation results (descriptions of these object
detectors can be found in \T~\ref{tab:model}).

We adopt two types of insertion schemes: \textit{random insertion} and
\textit{guided insertion}. Guided insertion works by randomly selecting one
existing object from the background image and inserting extra objects
\textit{close} to it. As shown in \F~\ref{fig:guided-insertion}, after randomly
selecting one elephant on the background image and inserting the ``bird'' image,
we create a \textcolor{blue}{blue} region that is \textit{symmetrical} and
\textit{centered} on the larger ``elephant.'' We randomly select locations
within the \textcolor{blue}{blue} region as the centroid of the ``bird.'' It is
easy to see that our sampling guarantees that the ``bird'' will not overlap with
the larger ``elephant.'' Moreover, overlapping with any other objects is
\textit{not} allowed either; whenever the ``bird'' is sampled over existing
objects in $i$ we discard the synthetic image and resample.

In contrast, the random insertion scheme implements a simple strategy in which
object $o$ is placed randomly on the background. Again, we disallow overlapping
of $o$ with existing objects and resample whenever overlapping occurs.
Additionally, for each background image $i$ with $N$ existing objects, we
perform $10 \times N$ guided or random insertions. We report the erroneous
object detector behaviors found with respect to the different setups as follows:

\begin{table}[H]
	\centering
	\scriptsize
  \resizebox{1.00\linewidth}{!}{
  \begin{tabular}{c|c|c|c}
	\hline
  \textbf{Object Detectors (see} & \textbf{\#Errors Found By} & \textbf{\#Errors Found By} & \textbf{\#Synthetic} \\
  \textbf{\T~\ref{tab:model} for introductions)}& \textbf{Random Insertion} & \textbf{Guided Insertion} & \textbf{Images} \\
	\hline
	 \textbf{Amazon Rekognintion}     & 232 & 432  & 2,270 \\
	 \textbf{Microsoft Azure Vision}  &  76 & 167  & 1,190 \\
	 \textbf{IBM Vision}              &  39 &  86  &  850  \\
	 \textbf{Google AutoML Vision}    & 365 & 461  & 1,570 \\
	\hline
	 \textbf{SSD Mobilenet}    &  45 & 112 & 1,290 \\
	 \textbf{SSD Inception}    & 185 & 310 & 1,510 \\
	 \textbf{RCNN Resnet}      & 218 & 381 & 2,980  \\
	 \textbf{RCNN Inception}   & 294 & 580 & 3,190  \\
	\hline
	 \textbf{Total} & 1,454 & 2,529 & 14,850 \\
	\hline
\end{tabular}
}
\end{table}

\noindent The object detectors identify different numbers of objects for each
image, and therefore we synthesize different numbers of images for testing. The
results show that the guided setting notably outperforms the first setting. This
is consistent with our intuition; by inserting images near the local region of
existing objects, the inserted images may disturb the regions or grids used for
object recognition and thus cause failures in object detectors.\footnote{The
  implementation details of these commercial object detectors are not
  disclosed.} As a result, \tool\ is configured with the guided insertion
strategy.


\begin{figure}[!ht]
  \centering \includegraphics[width=0.90\linewidth]{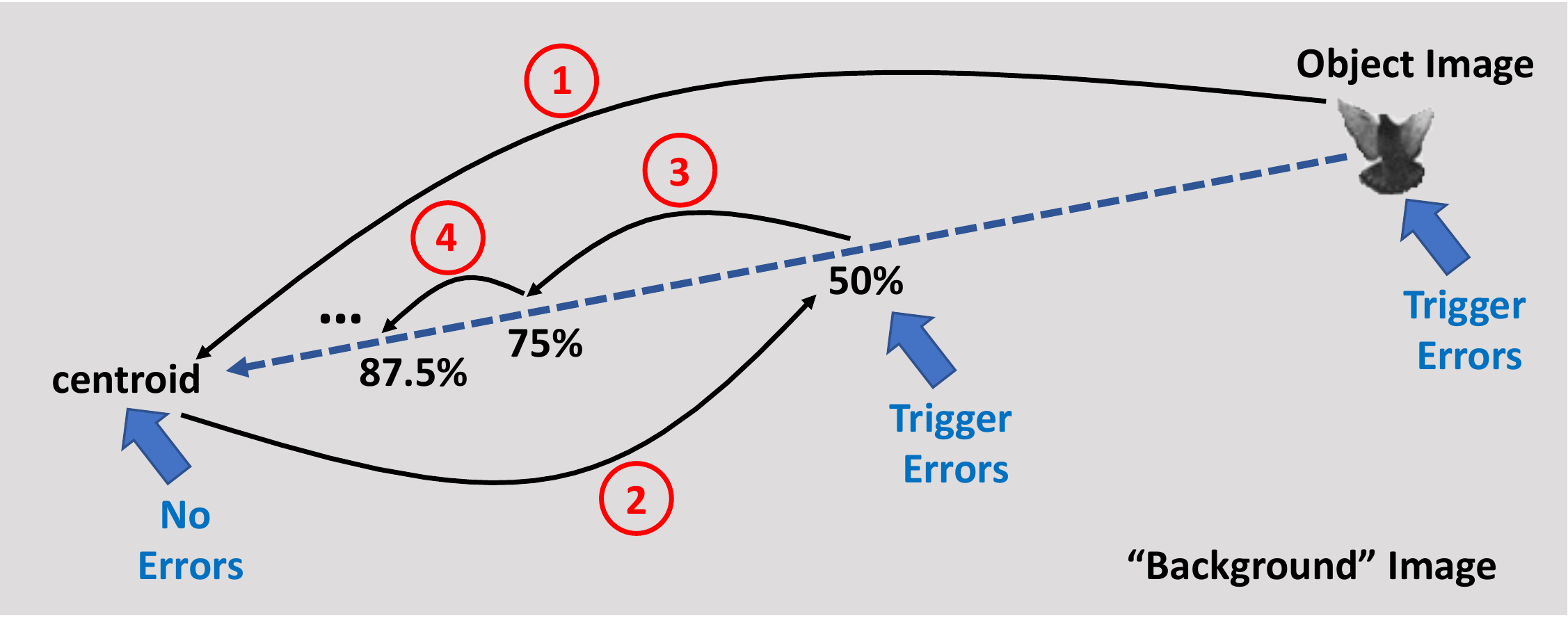}
  \caption{Move object instance image $o$ toward the centroid.}
  \label{fig:delta-workflow}
\end{figure}

\noindent \textbf{Augmenting Image Diversity.}~While the proposed techniques
provide practical guidelines on object insertion, ``guided insertion'' primarily
focuses on locations close to existing objects in the image and therefore may
miss opportunities for object insertion in other image regions. In this section,
we propose techniques to identify additional locations for object insertion,
with the goal of augmenting the diversity of synthetic images while still
maintaining their ``realism'' insofar as possible. To accomplish this, we first
compute the \textit{centroid} of the objects in the source image; then,
motivated by the use of delta debugging for conventional
software~\cite{zeller1999yesterday}, the inserted object is progressively
relocated toward the centroid while retaining the ability to cause prediction
errors.


The procedure is illustrated in \F~\ref{fig:delta-workflow}, where a
``delta-debugging''-style relocation scheme is implemented to explore locations
closest to the centroid. Starting from an insertion location found by the
guided-insertion strategy that can trigger object detector failures, we relocate
the inserted ``bird'' to the centroid of objects on the ``background'' image. If
no prediction error can be provoked regarding the newly synthetic image, we jump
back to the middle and recheck the object detector. In case this time prediction
errors do occur, we search forward until the ``bird'' becomes too close to 1)
the centroid; 2) the previous successful insertion (i.e., triggering prediction
errors) with the longest distance from the starting point; or 3) the starting
point itself. Again, for this step, we disallow any overlap between the inserted
object and existing objects on the background: whenever overlapping occurs, we
jump back as well.

It is worth mentioning that while the prototype implementation of \tool\ is
equipped to use ``centroid'' as the exploration destination, any locations could
be configured at this step to synthesize diverse and realistic images with
respect to user requirements.

\section{Implementation}
\label{sec:implementation}
\tool\ is implemented in Python in approximately 3,600 lines of code. As
mentioned earlier, the object extraction module of \tool\ is implemented by
extending the YOLACT~\cite{bolya2019yolact} instance segmentation framework. We
extended the framework by using the instance mask to crop the input image and
extract object instance images. The open-source YOLACT
implementation~\cite{yolact} is built with Pytorch (ver. 1.0.1), and contains
a model pretrained with a de facto object detection dataset,
COCO~\cite{lin2014microsoft}. This dataset contains objects with approximately
90 labels, and we use the pretrained model to perform instance segmentation. As
aforementioned, one desirable feature of YOLACT is that it performs instance
segmentation rapidly --- indeed, we performed all the instance segmentation tasks
using a single Nvidia GeForce GTX 1070 GPU. The overall processing time is
promising (for processing time evaluation, see \S~\ref{sec:evaluation}).

%
%
%

\begin{table}[!htbp]
	\centering
	\scriptsize
  \caption{Object detectors evaluated in this research. Due to the limited
    space, ``TensorFlow'' will be omitted from the model names. Also, TensorFlow
    faster RCNN Resnet and TensorFlow faster RCNN Inception Resnet will be
    referred as ``RCNN Resnet'' and ``RCNN Inception'', respectively.}
	\label{tab:model}
  \vspace*{-5pt}
	\resizebox{0.95\linewidth}{!}{
		\begin{tabular}{c|c|c}
			\hline
			\textbf{Object Detector Name} & \textbf{Speed} & \textbf{COCO mAP} \\
			\hline
			\textbf{Amazon Rekognintion API}~\cite{amazon2018rekognition} & fast & N/A \\
			\textbf{Google AutoML Vision API}~\cite{google2018api}        & fast & N/A \\
			\textbf{Microsoft Azure Vision API}~\cite{microsoft2019api}   & fast & N/A \\
			\textbf{IBM Vision API}~\cite{ibm2019api}                     & fast & N/A \\
      \hline
			\textbf{TensorFlow SSD Mobilenet}~\cite{howard2017mobilenets}  & fast & 21 \\
			\textbf{TensorFlow SSD Inception}~\cite{liu2016ssd}            & fast & 24 \\
			\textbf{TensorFlow faster RCNN Resnet}~\cite{ross2015fastrcnn} & medium & 32 \\
			\textbf{TensorFlow faster RCNN Inception Resnet}~\cite{szegedy2016rethinking}  & slow & 37 \\
			\hline
		\end{tabular}
	}
\end{table}

\section{Evaluation}
\label{sec:evaluation}

\begin{table*}[!htbp]
	\centering
	\scriptsize
  \caption{Result overview. Note that ``Processing Time'' includes the
    prediction time of object detectors.}
	\label{tab:results}
  \vspace*{-5pt}
	\resizebox{0.9\linewidth}{!}{
		\begin{tabular}{c|c|c|c|c|c}
			\hline
			\textbf{Object Detector} & \textbf{\#Synthetic Images} & \textbf{\#Detected Objects} & \textbf{\#Images Causing Detection Failures} & \textbf{Processing Time (Hours)} & \textbf{Total Cost (USD)} \\
			\hline
      \textbf{Amazon Rekognintion API}     & 38,939 & 3,750   & 6,060~~(15.6\%) & 11.9 & \$21.5 \\
      \textbf{Google AutoML Vision API}    & 18,655 & 1,801   & 2,738~~(14.7\%) & 13.0 & \$18.8 \\
      \textbf{Microsoft Azure Vision API}  & 20,453 & 1,985   & 2,494~~(12.2\%) & 3.2  & free \\
      \textbf{IBM Vision API}              & 13,280 & 1,290   & 1,515~~(11.4\%) & 2.2  & free \\
      \hline
      \textbf{SSD Mobilenet}    & 24,796 & 2,387   & 3,460~~(14.0\%) & 62.5 & \$53.4 \\
      \textbf{SSD Inception}    & 29,072 & 2,806   & 3,988~~(13.7\%) & 64.3 & \$54.9 \\
      \textbf{RCNN Resnet}      & 70,754 & 6,914   & 7,442~~(10.5\%) & 164.8 & \$140.7 \\
      \textbf{RCNN Inception}   & 76,257 & 7,349   & 10,648~~(14.0\%)& 290.8 & \$248.3 \\
			\hline
			\textbf{Total}                       & 292,206 & 28,282 & 38,345~~(13.1\%)& 612.7 & \$537.6 \\
			\hline
		\end{tabular}
	}
\end{table*}

\T~\ref{tab:model} lists the object detectors that we aim to test (the ``Speed'' and
``COCO mAP'' are mostly disclosed by Google~\cite{tensorflow2018sdk}). We use
four commercial object detection services provided by Amazon, Google, Microsoft,
and IBM for the
evaluation~\cite{amazon2018rekognition,google2018api,microsoft2019api,ibm2019api}.
We wrote Python scripts to interact with these remote services and retrieve the
prediction results (in JSON format). To the best of our knowledge, the object
detection models employed by these commercial services are not disclosed;
single-stage models are presumably employed given their prediction speed
(\S~\ref{subsec:object-detection}).
Google also supports directly deploying its TensorFlow object detection APIs on
Google Cloud~\cite{tensorflow2018sdk} and provides the flexibility to choose
different models pretrained on the COCO dataset~\cite{lin2014microsoft}. We
follow the official tutorial to setup TensorFlow object detection models on
Google Cloud~\cite{tensorflow2018sdk}, and from a total of five pretrained
models suggested in the accompanying tutorial, we choose four models, including
the RCNN Inception ResNet model~\cite{he2016deep}, which yields the best accuracy
but has the slowest speed. We also chose another RCNN model~\cite{wang2017fast}
and two SSD models~\cite{howard2017mobilenets,liu2016ssd} that exhibit medium
prediction speed and good accuracy. As mentioned in
\S~\ref{subsec:object-detection}, the two RCNN-based models have two-stage
region-based architectures, while the SSD models have single-stage architectures
that are much faster.

\subsection{Evaluation Overview}
\label{sec:eval-overview}
\T~\ref{tab:results} summarizes the evaluation results. To acquire these data, we
extracted object instances from 1,000 randomly selected images from the COCO
2017 image set~\cite{lin2014microsoft}. We then randomly selected 500 images
from the same dataset as background images. From the complete set of 1,000
images, \tool\ extracted a total of 5,843 object instances clustered with
respect to 79 different categories (person, dog, etc.). As previously mentioned
(\S~\ref{subsec:design-refinement}), the object refinement module of
\tool\ sorts object images with respect to their size and eliminates 90\% of the
small object images; the remaining 10\% of the object images are kept as
insertion candidates.

Given a background image $i$ containing $N$ objects, different object detectors
find different numbers of objects (the third column of \T~\ref{tab:results}
reports the total number of objects found by each detector). As discussed in
\S~\ref{subsec:design-insertion}, suppose that a detector finds $M$ objects in
$i$, then, \tool\ generates $10 \times M$ synthetic images following the
``guided-insertion'' strategy to test the detector. When a test image $i'$
triggers prediction errors, that image is used to generate additional diverse
test inputs following the ``delta-debugging''-style procedure
(\S~\ref{subsec:design-insertion}). The total number of images synthesized for 
each object detector is reported in the second column of
\T~\ref{tab:results}.

The number of images triggering prediction errors is reported in the fourth
column of \T~\ref{tab:results}. At least 10\% of the synthetic images triggered
erroneous predictions of the evaluated object detectors. \T~\ref{tab:results}
shows that object detection failures seem to be a general concern, regardless of
the underlying model. Moreover, when a model detects more objects in images, 
the number of images that can trigger failures is increased. We interpret this as
reasonable: recall that for an image of $M$ objects, our ``guided-insertion''
strategy generates $10 \times M$ synthetic images
(\S~\ref{subsec:design-insertion}).

\noindent \textbf{Processing Time.}~This part of the evaluation was conducted on a machine
equipped with an Intel i7-8700 CPU with 16 GB of RAM. The instance segmentation
module runs on a single Nvidia GeForce GTX 1070 GPU with CUDA 9.0.
\T~\ref{tab:results} also reports the processing time. In general, the
commercial APIs, particularly the Google and Microsoft services, require much
less time for prediction than that required by the TensorFlow pretrained
models. Although the implementation details of these remote services are not
disclosed, from the results, we can assume that the commercial remote services
presumably leverage highly optimized single-stage object detection models that
are faster but usually find fewer objects in images.

\noindent \textbf{Financial Cost.}~Enabled by modern cloud computing
infrastructures, all of these object detectors are designed as ``pay-as-you-go''
models: users are charged based on how many queries they send to the services
(for the first four services) or how many computing resources they use (for the
TensorFlow services). We report the amount of USD we are charged by these
services in \T~\ref{tab:results}. Due to the erroneous behavior, some of the
queries are indeed wasted. More importantly, given that commercial services have been
adopted in supporting critical computer vision applications (e.g., surveillance
cameras), we envision real-world scenarios where the prediction errors can cause
financial loss or fatal errors.

\subsection{Augment Diversity of Synthetic Images}
\label{sec:eval-augmentation}

\begin{figure}[!ht]
  \centering \includegraphics[width=0.98\linewidth]{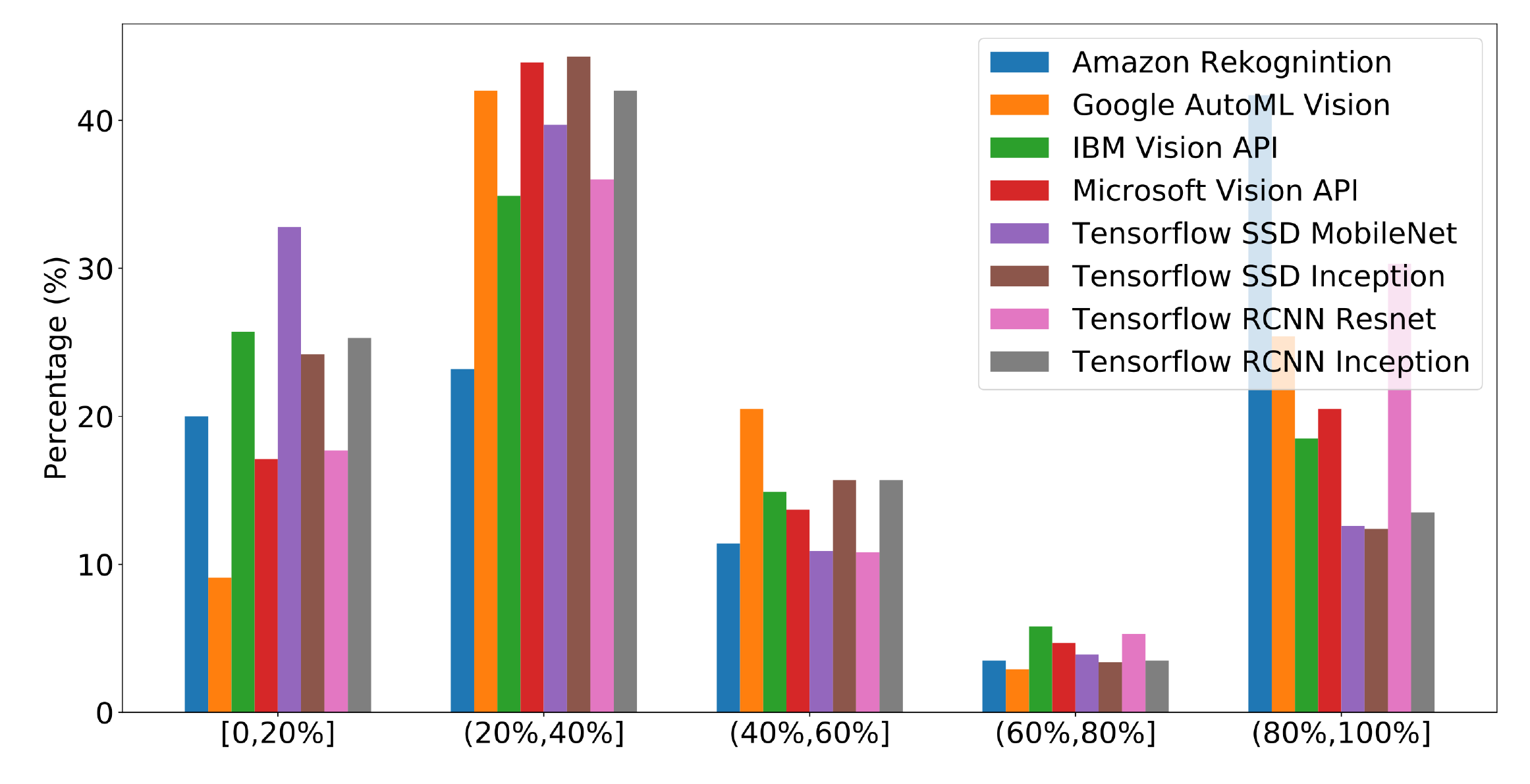}
  \vspace*{-5pt}
  \caption{Efficiency of synthetic image augmentation. Recall we leverage a
    ``delta-debugging''-style method to relocate the inserted objects toward
    the centroid (\F~\ref{fig:delta-workflow}). X-axis reports that how far the
    inserted object can proceed toward the centroid: 100\% indicates that the
    object is placed at the centroid.}
  \label{fig:augmentation}
\end{figure}

As discussed in \S~\ref{subsec:design-insertion}, enlightened by delta
debugging, we propose techniques to mutate synthetic images by progressively
moving an inserted object that triggers erroneous predictions toward the
centroid of objects in the background image. We preserve the realism of the
synthetic image at our best effort by placing the inserted object into a
realistic position, while augmenting the visual diversity of the synthetic
images.

\begin{table}[!ht]
  \centering
  \scriptsize
  \caption{Breakdown of images causing object detection failures. By adding the
    second and the last columns, we get the ``\#Images Causing Detection
    Failure'' column in \T~\ref{tab:results}.}
\label{tab:breakdown}
\vspace{-5pt}
  \resizebox{0.98\linewidth}{!}{
  \begin{tabular}{c|c|c|c}
  \hline
  \multirow{3}{*}{\textbf{Object Detector}} & \textbf{\#Synthetic}  & \textbf{\#Synthetic}   & \textbf{\#Unique Synthetic} \\
                                            & \textbf{Images With}      & \textbf{Images With}       & \textbf{Images With} \\
                                            & \textbf{Inserted Obj.} & \textbf{Relocated Obj.} & \textbf{Relocated Obj.} \\
  \hline
   \textbf{Amazon Rekognintion}           & 4,621  & 2,501& 1,439 \\
   \textbf{Google AutoML Vision}          & 2,093  & 853  & 645   \\
   \textbf{Microsoft Azure Vision}        & 1,891  & 742  & 603   \\
   \textbf{IBM Vision}                    & 1,135  & 481  & 380   \\
  \hline
   \textbf{SSD Mobilenet}      & 2,534 & 1,089 & 926   \\
   \textbf{SSD Inception}      & 2,976 & 1,192 & 1,012 \\
   \textbf{RCNN Resnet}        & 5,828 & 2,544 & 1,614 \\
   \textbf{RCNN Inception}     & 7,881 & 3,422 & 2,767 \\
  \hline
\end{tabular}
}
\end{table}

In this section, we study the efficiency of this augmentation method. We start
by reporting the breakdown of synthetic images causing object detection errors
in \T~\ref{tab:breakdown}. The second column reports the number of images
triggering prediction errors that are synthesized by inserting objects against
the background, while the third column reports the number of images triggering
prediction errors and are synthesized by relocating inserted objects toward the
centroid. Since the same object could be inserted at different positions on a
background image, and then reaching to the \textit{same centroid}, we also
measure the unique number of synthetic images at this step. As shown in the
\T~\ref{tab:breakdown}, the object relocation step successfully finds a large
number of images retaining the prediction errors of the object detectors. We
report that of a total of 28,959 synthetic images causing prediction errors,
9,386 (32.4\%) images are created via object relocation.
Moreover, we measure and report the average distance (in terms of percentage) by
which the inserted object can be relocated. Naturally, we consider arriving at
the centroid as 100\% and staying at the starting position as 0\%.
\F~\ref{fig:augmentation} reports the average distance data through barplots.
Note that \F~\ref{fig:augmentation} has excluded all the ``0\%'' cases, where
objects stay at the starting positions. As shown in the figure, on average
21.9\% of objects can be put at the centroid while retaining prediction errors,
and 40.2\% of object images are relocated at least 40\% of the distances.
Overall, we interpret the results as promising, illustrating that a considerable
number of synthetic images could be generated that retain prediction failures,
and also make the image visually more diverse.

\subsection{Naturalness of Synthetic Images}
\label{sec:naturalness}
In this section, we show that the synthetic images are still natural-looking.
While the ``naturalness'' of a synthetic image could be subjective to a certain
extent, as noted by existing research, natural images are deemed to have certain
statistical
regularities~\cite{huang1999statistics,mahendran2015understanding,mahendran2016visualizing}.
Therefore, following the convention of literatures in Computer
Vision~\cite{mahendran2016visualizing}, the ``naturalness'' of synthetic images
is measured by first computing a histogram of oriented gradients
(HOG~\cite{dalal2005histograms}) of both synthetic images and their
corresponding background images, and then computing the intersection of these
two HOGs. HOG is a popular metric extracting distribution (histograms) of
directions of gradients as ``features'' of an image. By summarizing the
magnitude of gradients, this metric captures abrupt intensity changes in the
image (object edges, object corners, etc.), and therefore is usually very
effective to comprehend high-level representations of images with multiple
objects. In contrast, pixel-level similarity metrics
(\S~\ref{subsec:design-refinement}) leveraged in \tool\ focus on single object
instance comparison, and are not applicable in this evaluation. Overall,
consistent with previous research~\cite{mahendran2016visualizing}, the
comparison output (i.e., HOG intersection), a value between 0 and 1, is used to
illustrate the naturalness of synthetic images.

\begin{table}[!ht]
  \centering
  \scriptsize
\caption{Naturalness evaluation w.r.t. average HOG intersection (higher value is
  better). We report evaluation results of both synthetic images with inserted
  objects (second column) and relocated objects (third column).}
  \vspace*{-5pt}
\label{tab:naturalness}
  \resizebox{0.88\linewidth}{!}{
  \begin{tabular}{c|c|c}
  \hline
  \multirow{2}{*}{\textbf{Object Detector}} & \textbf{Average HOG}  & \textbf{Average HOG} \\
                                   & \textbf{Intersection Rate (\%)}  & \textbf{Intersection Rate (\%)} \\
  \hline
   \textbf{Amazon Rekognintion}           & 98.9 & 98.8 \\
   \textbf{Google AutoML Vision}          & 98.1 & 98.0 \\
   \textbf{Microsoft Azure Vision}        & 98.8 & 98.7 \\
   \textbf{IBM Vision}                    & 98.5 & 98.5 \\
  \hline
   \textbf{SSD Mobilenet}      & 98.7 & 98.7 \\
   \textbf{SSD Inception}      & 98.6 & 98.6 \\
   \textbf{RCNN Resnet}        & 98.9 & 98.9 \\
   \textbf{RCNN Inception}     & 98.9 & 98.9 \\ 
  \hline
\end{tabular}
}
\end{table}

\T~\ref{tab:naturalness} reports HOG intersection rates (second column) by
comparing synthetic images with inserted objects and their corresponding
background images, and HOG intersection rates (third column) by comparing
synthetic images with relocated objects and their corresponding backgrounds.
Consistent with our intuition, all synthetic images have highly similar HOG
regularities with their corresponding backgrounds, and are deemed ``natural''
(in contrast, we report that the HOG intersection rate of two randomly-selected
images from the COCO dataset is less than 50.0\%). Also, while most synthetic
images with relocated objects have HOG intersections identical to those of the
synthetic images with inserted objects, there are three cases for which the
synthetic images with relocated objects exhibit a slightly lower rate.
Intuitively, relocation generates visually more diverse images and can
potentially lead to lower ``similarity'' comparing to their corresponding
background images.

\subsection{Prediction Failure-Aware Retraining}
\label{subsec:retraining}
To capitalize on the synthetic images that triggered prediction failures, we
show such synthetic images can be used to retrain models and substantially
improve the performances. To this end, we selected a popular autonomous-driving
dataset, Berkeley DeepDrive~\cite{bdd,yu2019bdd}, for this evaluation. This
dataset contains images that depict real-time driving experiences under
different weather conditions and at various times.
The experiments conducted in this section (i.e., model retraining) are executed
on a machine equipped with an Intel Xeon CPU E5-2680 with 256 GB of RAM and
eight Nvidia GeForce RTX 2080 GPUs.

We downloaded the SSD MobileNet object detection model pretrained by TensorFlow
and retrained the model (with Tensorflow ver. 1.14.0) by using 900 images
annotated with 10 common categories for traffic scenes from the DeepDrive
training set. We actually imitated how object detection models are customized
and used in practice; based on transfer learning~\cite{pan2009survey},
pretrained models are adapted to similar tasks by fine-tuning the model
parameters on a new dataset. At this step, we reuse the \textit{default
  configuration} shipped with the MobileNet pretrained model; the batch size is
48 which means the whole training set will be processed once within 19 steps. We
set up three retraining strategies (\textit{Config$_{1-3}$}) as follows:

\begin{itemize}
  \item We start by retraining the MobileNet model with 900 images for 200K
    steps (200K is the default setting in the model's configuration) and
    exporting the retrained model $m_0$. We also form a evaluation set by
    randomly selecting 100 images from the DeepDrive evaluation set.
  \item We then use \tool\ to generate new synthetic images from the 900 images
    and collect synthetic images that cause prediction failures of $m_0$. This
    step generates 18,707 images (denoted as $\mathcal{I^{*}}$) triggering
    prediction errors.
  \item \textit{Config$_1$}: starting from $m_0$, we resume retraining with the
    900 images for another 10K steps.
  \item \textit{Config$_2$}: starting from $m_0$, we extend the original
    training set of 900 images with 900 images randomly selected from
    $\mathcal{I^{*}}$, and resume the model retraining with these 1,800 images
    for another 10K steps. To label each synthetic image, the label of its
    corresponding ``background'' image is reused.
  \item We also use \tool\ to generate another set of 900 images (denoted as
    $\mathcal{I}$). We do \textit{not} check whether these images can trigger
    prediction failures or not.
  \item \textit{Config$_3$}: starting from $m_0$, we extend the training set of
    900 images with 900 images in $\mathcal{I}$, and resume the model retraining
    for another 10K steps. Again, to label each synthetic image, the label of
    its corresponding ``background'' is reused.
\end{itemize}
During the retraining of in total 210K steps, we measure the total loss and mAP
score regarding the evaluation set of 100 images.

\begin{figure}[!ht]
  \vspace*{-10pt}
  \centering
  \includegraphics[width=1.0\linewidth]{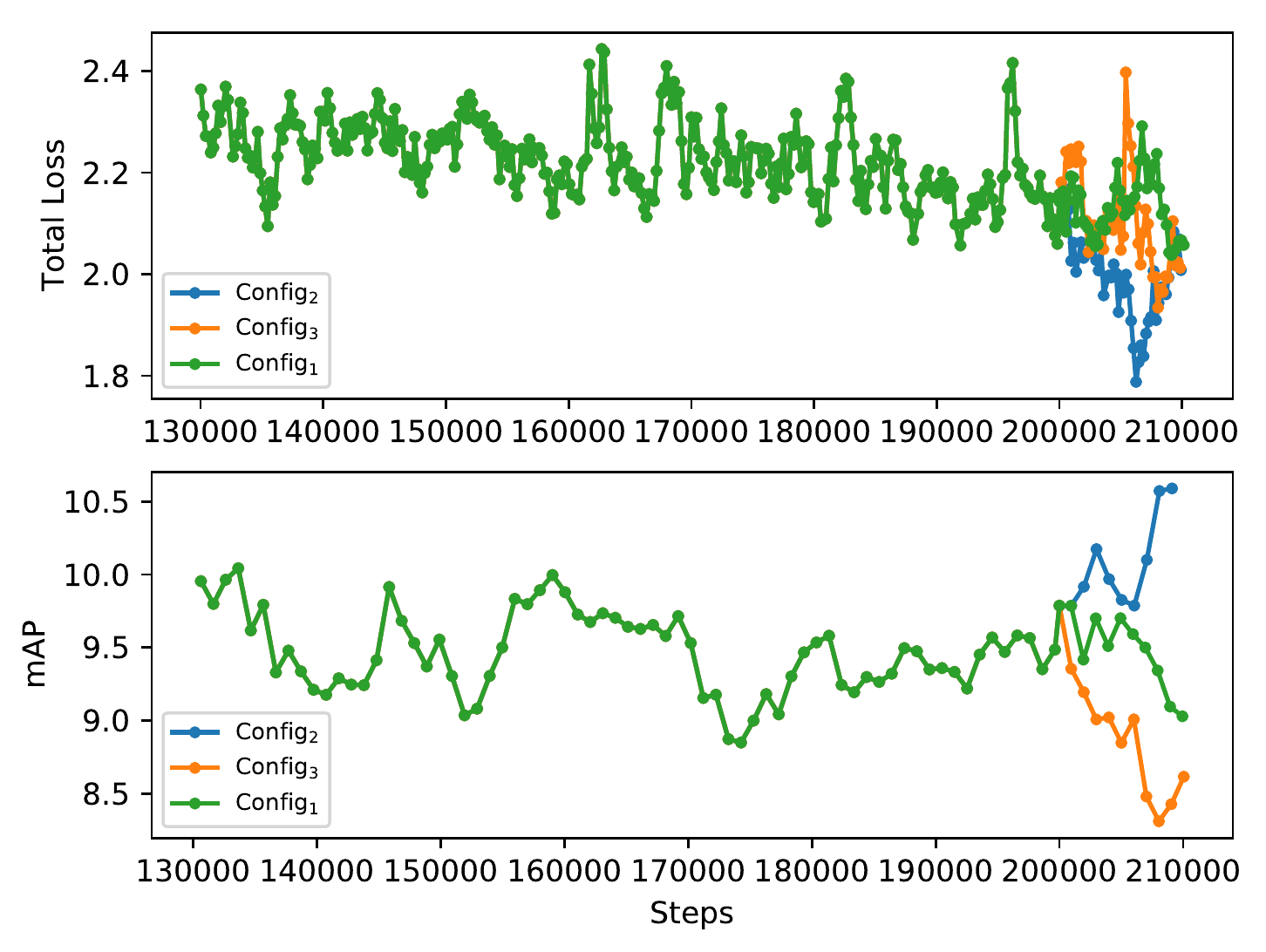}
  \vspace*{-15pt}
  \caption{Smoothed total loss (first diagram) and mAP scores (second diagram)
    during model retraining. Due to limited space, we show performance starting
    from 130K steps.}
  \label{fig:total-loss}
\end{figure}

As reported in \F~\ref{fig:total-loss}, for the last 10K steps,
\textit{Config$_1$} shows consistent trending compared to the first 200K steps.
\textit{Config$_2$} outperforms the other two by having lower total loss.
Moreover, the mAP score of \textit{Config$_2$} clearly outperforms those of the
other configurations. \textit{Config$_3$} exhibits a slightly better total loss
decrease than that of \textit{Config$_1$}, but yields an even lower mAP score
(which may due to overfitting). We report that the average mAP scores of the
three configurations within the last 10K steps are as follows:

\vspace{-8pt}
\begin{table}[H]
	\centering
	\scriptsize
  \resizebox{0.5\linewidth}{!}{
  \begin{tabular}{|c|c|c|c|}
	\hline
   & \textit{Config$_{1}$} & \textit{Config$_{2}$} & \textit{Config$_{3}$} \\
	\hline
  \textbf{mAP} & 9.3 & 10.5 & 8.6 \\
	\hline
\end{tabular}
}
\end{table}
\vspace{-8pt}

As the table shows, model performance is increased by retraining with the
synthetic images of the prediction errors. Note that according to object
detection surveys (e.g., Table Two in~\cite{zhong2018comparison} and Table Seven
in~\cite{liu2018survey}), one point mAP score increase is significant.
Overall, we interpret the evaluation result as promising: the failure-aware
retraining demonstrated in this section sheds light on practical usages of the
model prediction errors found by \tool\ and provides promising directions to
improve model accuracy.

We acknowledge that the evaluation, while being \textit{fair}, may not
illustrate the \textit{best practice} to promote model performance;
\F~\ref{fig:total-loss} indicates that the ``sweet spot'' might be around 8K
steps of retraining (where the mAP score is approximately 10.7). Overall, we
consider that providing guidelines of best practice is beyond the scope of this
research, but the reported results have illustrated the potential. Additionally,
images are synthesized from the existing training set; in other words, we do
\textit{not} need new real images. Overall, ``failure-aware'' retraining is
\textit{orthogonal} to standard model retraining techniques and can potentially
be orchestrated together.

\section{Discussion and Future Work}
\label{sec:discussion}
In this paper, we presented the design, implementation and evaluation of \tool,
a systematic workflow for automatically testing the erroneous behaviors of
object detection systems. The proposed techniques can be adopted to promote
object detector training and to motivate this emerging line of research. In this
section, we present a discussion and several potentially promising directions
for future research.

\noindent \textbf{Comparing to works in the CV Community.}~Parallel to SE
community's efforts on testing deep learning systems, the CV community generates
synthetic inputs by mutating real images to train deep neural networks. We
compare and illustrate the novelty of \tool\ with related CV research along
several aspects:

\begin{itemize}
  \item \textit{Blackbox vs. Whitebox}: most existing CV research considers a
    ``white-box'' setting (e.g., \cite{rozantsev2014on,peng2014exploring}). Such
    efforts either require a deep understanding of the model structure to
    adaptively synthesize inputs~\cite{rozantsev2014on}, or use the hidden
    layers of the neural network model to directly guide input
    synthesis~\cite{peng2014exploring}. In contrast, as aforementioned, our work
    considers the blackbox setting for software testing and introduces \tool\ to
    effectively test commercial off-the-shelf object detection models.
  \item \textit{Training with synthetic inputs vs. Re-training with
    bug-triggering synthetic inputs}: To our knowledge, all related CV research
    directly uses synthetic inputs to train the model. In contrast, MetaOD
    suggests a novel failure-aware model retraining scheme (cf.
    \S~\ref{subsec:retraining}) to effectively improve model accuracy, which
    suggests the interesting future work to continue testing the ``re-trained''
    model, i.e., the whole process would loop to iteratively re-train the model.
    We expect the model accuracy to further improve until reaching saturation.
    Note also that as evaluated in \F~\ref{fig:total-loss}, model re-training
    with arbitrarily generated synthetic images may lead to decreased model
    performance, while re-training with bug-triggering synthetic inputs leads to
    significantly improved model accuracy.
 \item \textit{Fine-grained modeling/tuning on the synthetic image for training
   vs. generic framework to synthesize images for testing}: As already discussed
   in related work, existing CV research mostly performs heavyweight,
   fine-grained (statistical) modeling to generate synthetic images (e.g.,
   \cite{Black03anovel}), where trajectory is particularly considered to
   synthetic images and train surveillance tracking systems. As a result, these
   techniques usually focus on specific application domains, and leverage domain
   knowledge to fine-tune and optimize the synthetic images. In contrast, our
   goal is to design a general framework to efficiently generate a large amount
   of quality inputs for testing. MetaOD is, in general, agnostic of image
   ``semantics'' (except labels of existing objects in the image) and therefore
   can be more efficient and robust.
\end{itemize}

\noindent \textbf{Novelty Comparing to DL testing work in SE community.}~As
discussed in \S~\ref{sec:related}, most existing testing work in the SE
community focuses on image classification models (e.g.,
\cite{pei2017deepxplore,tian2018deeptest}), or the underlying infrastructures of
TensorFlow/PyTorch (e.g.,
\cite{zhang2018empirical,pham2019cradle,kim2019guiding,dutta2018testing}). To
our knowledge, no prior work focuses on designing a general, effective pipeline
to test object detection models, another class of fundamental models used in
many real-world critical applications. \tool\ mutates and observes the detection
of individual objects in an image, while existing research on testing image
classification performs whole image-wise mutations (e.g., adding foggy and rainy
conditions in the image). Some of these transformations are not applicable to
our scenario (as discussed in \S~\ref{subsec:application-scope}), and our
approach is generally orthogonal to these whole image-wise mutations.

Our work also demonstrates the feasibility of ``failure-driven retraining''
(\S~\ref{subsec:retraining}) with notably improved model performance. This
evaluation addresses a typical concern in the SE community on ``how to use
findings of DNN testing'', which is not well explored by previous testing work
in this area.

\noindent \textbf{Boost Object Detector Testing with Generative Adversarial
  Network (GAN).}~Careful readers may wonder about the feasibility of
synthesizing images with a GAN~\cite{goodfellow2014generative}. While in
principle this technique is legitimate and potentially promising, in practice,
we argue that a GAN cannot provide a principled guarantee of the generated
images. It is commonly acknowledged that with a GAN, the blending of objects is
usually fuzzy and blurry. Additionally, GAN are usually very difficult to train,
especially in our usage scenarios, where a unified solution is expected to
produce arbitrary images for testing. We leave it to future work to explore
practical solutions for using GANs in object detection testing systems. The
software engineering community has proposed techniques for generating mutated
programs to test compilers that inserting or remove arbitrary statements from a
test-case program~\cite{le2014emi}.Thus, a similar question is whether it is
true that could simply removing existing objects from test-case images to check
whether the mutated images lead to inconsistent prediction reasons; intuitively,
object detection results on the mutated image should be consistent with the
input image. Although this could also be a straightforward and promising
direction to explore, removing an object from the image will leave a ``blank''
space, possibly making the output image unrealistic. We note that some recent
research works have trained a GAN model to ``inpaint'' the empty space with
hints from the surrounding context~\cite{pathak2016context}. As a next step, it
would be interesting to test the object detectors by removing arbitrary objects
and leveraging an inpainting model to reconstruct the removed areas.

\noindent \textbf{Boost Object Detector Testing with Deep Reinforcement
  Learning.}~As explained in \S~\ref{subsec:design-insertion}, one key challenge
in testing object detection systems is to determine a reasonable position to
insert additional object images. Despite the promising experimental results
revealed by using the present workflow, in future work, it would also be
interesting to explore the feasibility of leveraging reinforcement learning to
infer the optimal insertion location.

Given a test-input image, a reinforcement learning agent could be trained to
determine the optimal location for inserting an object image. At each step, the
agent would find a location $(x, y)$ on the image to paste the object and would
agent receive a reward from the remote object detector when the synthetic image
triggers an erroneous behavior. A reinforcement learning agent is trained with
two types of inputs during each step—the reward and a state; in our setting, the
``state'' is the synthetic image, which means that convolutional layers would be
needed to directly seek to learn from the high-dimensional inputs.

The action space in our setting is quite large (conceptually every location on
the test-input image forms a potential location for pasting objects), which can
cause major difficulty during training. Again, as future work, it would be
interesting to model object insertion as a ``maze escape'' puzzle; instead of
determining an arbitrary insertion location, we only decide in which of four
directions to move from the previously inserted location. This scheme reduces
the action space to four possibilities.

% ...

\noindent \textbf{Usage Scenario of \tool.}~Most of the testing work to-date
targeting computer vision models aims to find model prediction errors during the
system testing stage. \tool\ can also be used during that stage. Moreover, we
consider \S~\ref{subsec:retraining} has shed light on the interesting
possibilities to integrate MetaOD during the system development (model training)
stage. As mentioned in Novelty Comparing to works in the CV community of our
rebuttal, the overall workflow can form a loop, where \tool\ continues
identifying and adding error-triggering images into the training data set to
re-train the model. Our results suggest that such a workflow can improve model
performance.

As discussed in the last paragraph of \S~\ref{subsec:retraining}, while
proposing the best practice at this step to launch ``Prediction Failure-Aware
Retraining'' is out of the scope for our current work, it would be highly
interesting to further explore beneficial use cases for such ``Prediction
Failure-Aware Retraining'' scheme and boost the model training stage. We leave
it as an interesting, novel direction for future research.

\noindent \textbf{Realism/Naturalness of Synthesized Images and Its
  Effectiveness of Metamorphic Testing.}~ We consider that
``realism/naturalness'' does not have direct influence on the effectiveness of
metamorphic testing. Randomly mutating pixels to generate ``fuzzy'' and
``unreal'' images as the test inputs, which are challenging for human eyes to
detect objects, could also be used for stress-testing object detectors.

However, we consider ``realism'' is beneficial in this research because:

\begin{itemize}
  \item As discussed in Section 3.3 ``Application Scope'', we are not testing
    extreme cases to stress object detectors (not like a typical fuzz testing
    setting). Stress testing of object detectors would be different, and in
    general, we believe that our community is not quite there yet.
  \item Synthesizing more ``realistic'' images facilitate the practical usage of
    MetaOD. As noted in the \S~\ref{sec:approach-overview}, we aim to also
    augment the realism such that the synthesized images triggering erroneous
    predictions can mostly reveal practical defects that can likely cause
    confusion during daily usage of object detectors.
\item It is more reasonable to use these \textit{realistic} and error-triggering
  images to retrain object detection models (see \S~\ref{subsec:retraining})
  than using arbitrary error-triggering inputs.
\end{itemize}

\noindent \textbf{Definition of Image Naturalness.}~As discussed in
\S~\ref{subsec:design-insertion}, it is generally challenging, if not
impossible, to understand the ``semantics'' of each object instance and
accordingly perform fine-grained rotation and insertion to be fully consistent
with the background. The computer vision community is exploring methods to
address this challenge, and to our knowledge, the state-of-the-art uses
heavyweight statistical methods and only applies them to mutate human gestures
in the images of indoor scenes~\cite{qi2018human}. We do not tackle this
challenge to comprehend the fine-grained meaning/gesture of each object (e.g., a
car). Rather, we introduce a general, practical pipeline to effectively pinpoint
erroneous predictions given arbitrary images.

\section{Related Work}
\label{sec:related}
\vspace{-3pt}
\noindent \textbf{Testing of Deep Learning Systems.}~
Testing techniques for conventional software have been recently applied for deep
learning systems, including fuzz
testing~\cite{odena2018tensorfuzz,xie2018coverage}, mutation
testing~\cite{wang2019adversarial,ma2018deepmutation}, metamorphic
testing~\cite{dwarakanath2018identifying,zhang2018deeproad}, and also symbolic
execution~\cite{sun2018concolic,sun2019deep,gopinath2018symbolic}. The majority
of existing work focuses on image classification and its adoption on autonomous
driving systems~\cite{zhang2018deeproad,tian2018deeptest,pei2017deepxplore}.
Udeshi et al. tested the fairness of deep learning
systems~\cite{udeshi2018automated}. It is worth noting that previous work on
testing deep learning systems often adopts ``differential testing''
schemes~\cite{tian2018deeptest,pei2017deepxplore}; however, object detection
models can usually recognize different number of objects from an image (due to
the model capability), leading to the general challenge for cross checking.
In contrast, this research adopts metamorphic testing as an effective and
adaptive testing strategy to reveal defects in these commercial object
detectors.
Regarding the testing oracle selection, neuron
coverage~\cite{pei2017deepxplore,tian2018deeptest} and other finer-grained
coverage metrics have been proposed~\cite{ma2018deepgauge}.
Also, in addition to the deep learning models, the underlying infrastructures
(e.g., TensorFlow~\cite{abadi2016tensor} and PyTorch~\cite{paszke2017automatic})
have also been tested to find implementation
bugs~\cite{zhang2018empirical,pham2019cradle,kim2019guiding,dutta2018testing}.

\noindent \textbf{Data Augmentation for Object Detection Model Training.}~In
parallel to the SE community's promising progress in testing deep learning
systems, data augmentation, which generates synthetic inputs by mutating real
images, has become an important technique to train deep neural networks. To
train image analysis deep model (e.g., for object detection), the proposed
augmentation methods vary from geometrical transformations such as horizontal
flipping to color perturbations to adding noise to an image (e.g., mimic severe
weather
conditions)~\cite{gaidon2016virtual,dwibedi2017cut,tremblay2018training,frid2018synthetic,hinterstoisser2018pre,zheng2017unlabeled,prakash2018structured,georgakis2017synthesizing}.
In fact, it has been shown that the model accuracy can usually be improved by
taking such synthetic images into the training
set~\cite{georgakis2017synthesizing}.

Most existing work prioritizes \textit{local} rather than global consistency
when augmenting images. For instance, while some approaches insert random objects
into training images, these studies have focused more on the realism of the
inserted objects than on the context surrounding. Many synthetic images are
unrealistic from a global point of view, such as putting a car on the
table~\cite{tremblay2018training}. A few studies have leveraged heavyweight
statistics methods to infer a ``realistic'' location for object insertion; they
assumes a ``white-box'' setting and can handle only a few domain-specific
scenes~\cite{qi2018human}. In contrast, the present work proposes a lightweight
and systematic new focus to promote the synthetic images by considering both
local and global realism. We take a ``black-box'' setting that facilitates the
testing of commercial remote object detection models. More importantly, our
testing focus enables a unique ``failure-aware'' model retraining scheme
(\S~\ref{subsec:retraining}), which effectively improves the model accuracy.


\vspace{-2pt}
\section{Conclusion}
\label{sec:conclusion}
\vspace{-2pt}
Object detectors powered by deep neural networks have been commonly used
in real-world scenarios. This paper has introduced a novel metamorphic testing
approach toward reliable object detectors.
%
%
Evaluation results are promising --- \tool\ can find thousands of prediction errors,
and generated synthetic images can be used for retraining and substantially
improve model accuracy.

\bibliographystyle{ACM-Reference-Format}
\bibliography{bib/ref,bib/cv,bib/testing-cv}

\end{document}